\documentclass[acmsmall]{acmart}

\AtBeginDocument{%
  \providecommand\BibTeX{{%
    \normalfont B\kern-0.5em{\scshape i\kern-0.25em b}\kern-0.8em\TeX}}}

\setcopyright{acmcopyright}
\copyrightyear{2018}
\acmYear{2018}
\acmDOI{XXXXXXX.XXXXXXX}

\acmConference[ACM CSUR '24]{Make sure to enter the correct
  conference title from your rights confirmation emai}{June 03--05,
  2018}{Woodstock, NY}

\acmISBN{978-1-4503-XXXX-X/18/06}

\usepackage{caption}
\usepackage{subcaption}
\usepackage{longtable}
\usepackage{pdflscape}
\usepackage{multicol}
\usepackage{multirow}
\usepackage{xtab}
\usepackage{array}
\usepackage{enumerate}
\usepackage{enumitem}
\newcommand{\red}[1]{\textcolor{black}{#1}}

\usepackage{multirow}
\begin{document}

\title[A Survey on recent trends in Robotic Prosthetics]{A Survey on Robotic Prosthetics: Neuroprosthetics, Soft Actuators, and Control Strategies}

\author{Jyothish K. J.}
\email{jyothishk.j@niser.ac.in}
\orcid{0000-0001-7731-3281}
\author{Subhankar Mishra}
\email{smishra@niser.ac.in}
\orcid{0000-0002-9910-7291}
\affiliation{%
  \institution{National Institute of Science Education and Research, Bhubaneswar - An OCC of HBNI.}
  \streetaddress{Jatni}
  \city{Khordha}
  \state{Odisha}
  \country{India}
  \postcode{752050}
}

\renewcommand{\shortauthors}{Jyothish and Subhankar}

\begin{abstract}
  The field of robotics is a quickly evolving feat of technology that accepts contributions from various genres of science. Neuroscience, Physiology, Chemistry, Material science, Computer science, and the wide umbrella of mechatronics have all simultaneously contributed to many innovations in the prosthetic applications of robotics. This review begins with a discussion of the scope of the term robotic prosthetics and discusses the evolving domain of Neuroprosthetics. The discussion is then constrained to focus on various actuation and control strategies for robotic prosthetic limbs. This review discusses various soft robotic actuators such as EAP, SMA, FFA, etc., and the merits of such actuators over conventional hard robotic actuators. Options in control strategies for robotic prosthetics, that are in various states of research and development, are reviewed. This paper concludes the discussion with an analysis regarding the prospective direction in which this field of robotic prosthetics is evolving in terms of actuation, control, and other features relevant to artificial limbs. This paper intends to review some of the emerging research and development trends in the field of robotic prosthetics and summarize many tangents that are represented under this broad domain in an approachable manner.
\end{abstract}

\setcopyright{acmlicensed}
\acmJournal{CSUR}
\acmYear{2024} \acmVolume{1} \acmNumber{1} \acmArticle{1} \acmMonth{1}\acmDOI{10.1145/3648355}




\received{19 October 2022}
\received[revised]{23 October 2023}
\received[accepted]{31 January 2024}

\maketitle

\section{Introduction}
In the last few decades, we have grown to see robots as viable tools and, in many cases, as able extensions to our bodily appendages. We are now able to execute tasks that were otherwise beyond human capabilities with precision and consistency that’s unseen before. Robotic technology has seen a significant number of applications in the field of medicine in recent years.  The technology itself has been becoming increasingly feasible, and the acceptance of robots for assistance even in intricate tasks such as surgery has grown significantly. Today, we can do surgeries that require microscopic precision with a commendable success rate; surgeons can operate on organs as delicate as the eyes, brain, and heart with repeatable and non-fatiguing precision. Not only surgery, assistance for the elderly, childcare, medical diagnosis, mobility assistance for the disabled, and of course prosthetics are just some of the applications that are vastly benefitted by technological advancements in the field of robotics. To understand their contribution better,  A popular classification of medical robots, as given by Cianchetti et al. \cite{BiomedSoftRob} is given below:
\begin{itemize}
\item Medical devices: Robotic devices used in surgery, diagnosis, and drug delivery.
\item Assistive robots: Rehabilitation and wearable robots.
\item Prosthetic devices: Robots that mimic human body parts such as prosthetic limbs, body-part simulators and artificial organs. 
\end{itemize}
 Medical robots range from huge robotic surgical systems such as daVinci \cite{davincireview} to catheter robots to tiny pill cameras \cite{pillcamera} and microchip type robots \cite{Neuralink}. Active research is underway in the field of nanorobotics which feature molecular level components measuring in the range of nanometers (10\textsuperscript{-9})\cite{BioMechatronics}. daVinci and other such robots represent a telerobotic system where one or more patient-side slave robots are operated by a qualified human surgeon with the help of a master unit that captures the surgeon’s movement through highly precise joystick-like inputs. These techniques allow for Minimally Invasive Surgery (MIS) for a variety of cardiac, gastrointestinal, urological, and other surgical procedures. An optimal telepresence surgical system (or surgical telerobotic system) would provide the surgeon the feeling of having their hands virtually inside the patient’s body \cite{Telepresence} while actually operating with small-scale surgical tools that are mounted on the end of the patient side of the robot robots \cite{SurgRob}. A discussion about these haptic systems in some detail is covered in the Secion \ref{sec:ermr} of this paper. Specialty surgical robots that are designed for operating on specific parts of the human body are also prominent \cite{Robodoc}.
Modern advancements in AI  and data science are routing further refinement in robot-assisted surgeries, for example, the cloud-connected surgical robot daVinci can browse through large sets of open MRI data of patients with similar condition \cite{davinciCloud} for procedural refinement, which is a commendable feat. Further, these surgical robots also facilitate linking with more traditional diagnostic tools such as CT scans, X-rays, etc. during the surgery for precise guidance of the operating apparatus.  Likewise, the prominence of robotic solutions in patient care and medical assistance is also noteworthy. Robots are becoming more and more mainstream when it comes to rehabilitation \cite{physiotherapyrobot}, mobility assistance \cite{ReWalk}, and even in the area of personal telepresence related to clinical care \cite{rp-vita}. Robotic prosthetic devices are already in the mainstream of medical sciences. pacemakers \cite{pacemaker}, insulin systems \cite{insuline}, and robotic limbs \cite{RoboLeg} are just a few examples of this pioneering feat of technological advancement in robot-assisted medical sciences. \\

It is clearly evident that medical sciences are hugely benefiting from presently, and are going to be closely reliant on robotics in the near future. Hard robotics-based devices that utilize motors and gears for motion generation have been a norm in this field.  However, such traditional devices have their limitations.  Lack of organic behavior is the most prominent flaw when it comes to hard robotics-based devices.  This becomes evident in their lack of compliance when affected by factors such as accidental collisions with people/objects, falling down, etc.  Moreover, traditional forms of motion generation such as motors are far from ideal for such applications involving humans due to a  multitude of reasons,  which will be discussed ahead. The use of soft material for robotic applications is seen in increasing amounts in various fields of medical sciences, especially in surgery.  When it comes to motion assistive systems such as a prosthetic limb, the set of challenges is similar, i.e., to guide mass displacement by providing the necessary force.  However, since these devices are in close proximity to the user, they must be safe, effective, and able to employ force without endangering the person.

This paper comprehensively analyzes the scientific advancements in the fields of Medicine, Chemistry, and Engineering that may carve the way to prosthetics of the future. The potential withheld by smart materials, composites, and their use in fabricating artificial muscles, that can function similarly to their natural counterpart, is one area of focus throughout this paper. The second area of focus is to understand and evaluate the options available for control systems for such robotic prosthetics. Various available options from the current technology will be explored along with studying the various pioneering research in the fields of robotic prosthetics, neuroprosthetics, artificial muscles, control systems, and other parallel trends such as haptics, kinesthetics, proprioception, etc. 

This paper condenses information from various domains and is inspired by literature from various subsections. Contributions of this paper along with some related work from literature have been summarized below:

\begin{itemize}
    \item Scope of prosthetics and robotic prosthetics \cite{NTReview}, \cite{ProsDefn}, \cite{RobProsRev}
    \item Overview of Neuroprosthetics as \red{grounds for various sensory rehabilitation and prosthetic control applications} \cite{NEuroprosReview}, \cite{ProsBrain}, \cite{BCIApp}, \cite{Ch16ProsBrain}, \cite{Ch9NPLeg}, \cite{HumRobInt}, \cite{NEuroprosReview}, \cite{NIReview}
    \item \red{Sensory rehabilitation applications for restoration of vision, hearing, and touch. \cite{VisProsAssDev},  \cite{HearingAid}, \cite{BION}} 
    \item \red{Preference of soft robotics over} hard robotics for prosthetic applications
    \item Review of the field of soft robotic actuators \cite{BiomedSoftRob}, \cite{SMABook}, \cite{SelfHealReview}, \cite{MaterJamTypes}, \cite{Electrorheology} 
    \item Electroactive polymers \cite{EAPLatRev}, \cite{CurrSilic}, \cite{EAPChemRev}
    \item Electrorheological and Magnetorheological actuators for haptics \cite{MRFeasibility}, \cite{RevMRER}
    \item Overview of various control strategies for robotic prosthetic limbs  \cite{BCIRobpro}, \cite{HMIReview}, \cite{FESArm} 
    \item EMG and other residual muscle-based control strategies for robotic prosthetic limbs \cite{ReviewControl}, \cite{EMGNonInv}
    \item \red{Interfacing prosthetic limbs electrically with nervous system \cite{Ch9NPLeg}, \cite{Neuralink}}
    \item \red{Artificial skin \cite{ProstheticSkin}, \cite{ReviewFlexTactHIS}, \cite{TouchSensingReview}}
    \item Sensory restoration (Touch and proprioception etc.) in robotic prosthetic limbs, Aesthetics, complications of limb amputation and preference to passive prosthetic limb over the active counterparts \cite{Aesthetic}, \cite{Proprioception}, \cite{PiezoTouch}, \cite{Eskin}, \cite{NeuroImpl}, \cite{TimeDisVibro}, \cite{Electrocutaneous}
\end{itemize}

 The next section (Sec. \ref{sec:prosthetics}), looks at the diversity of mechatronic systems that come under the domain of robotic prosthetics. \red{This review looks at the field of neuroprosthetics in some detail as the principles covered here will power the discussion of control strategies for robotic prosthetics and advanced sensory integration between the prosthetic and human.} \red{In the following section \ref{sec:motor}, }The discussion will be narrowed down to the field of robotic prosthetic limbs and continue looking at the current state of technology in robotic prosthetic limbs including commercial availability, feasibility, and the issues with conventional hard robotic actuators when it comes to prosthetic applications. A discussion about various soft robotic actuators and the benefits offered by them for biomedical applications will be presented. The next section will discuss popular control strategies for robotic prosthetics and will also talk about the potential direction that the field is headed towards. \red{The paper will then discuss parallel trends along its main discussion including artificial skin, proprioception, etc}. Finally, the paper concludes with some impressions about the future of robotic prosthetics. The supplementary material to this paper contains some illustrations to enhance the reader's understanding of the scope of neuroprosthetics, application sites of visual and auditory prosthetics, the working mechanism of material jamming actuators, and tendon-driven mechanisms that are covered as a part of our discussion on soft actuators in section 3 and workflow of EMG based control machinery for robotic prosthetics.

\section{Prosthetics:}
\label{sec:prosthetics}

Prostheses are externally applied devices and items that help persons with physical disabilities or functional restrictions to improve their function and boost their ability to live productive, healthy, and independent lives.
The term “Prosthesis” originates from the amalgamation of Greek words “Pro-” which means “Instead of” and “Thesis” which means “placing”. The term means the replacement of whole or part of a diseased or damaged bodily organ by an artificial device \cite{ProsDefn}. While the term prosthetics is usually interpreted under the purview of movement-related appendages, the scope of prostheses is vast. Obviously an artificial hand \cite{RobProsHand} \cite{HandDext} or leg \cite{RobLegPros} \cite{OpenLeg} used by an amputee is called a prosthetic, but much less obvious instruments such as an Intrauterine Contraceptive Device or  Urinary Stents are also valid prosthetic devices. Dental Implants, Orthopedic Implants, Exoskeletons\cite{RobProsRev}, Vascular Stents, Artificial Heart Valves, Hernia Mesh, Breast and Penile implants, Hearing Aid, etc. are just some of the objects healing people in our society by treating their impairments. As per WHO \cite{WHOdef}, Assistive Products (Classified under Assistive Technologies) can be further divided into Hearing, Vision, Cognition, Communication, Mobility, and Environment assistive products. Prosthetics are defined as a subset of Mobility Assistive Products along with orthotics, wheelchairs, etc. This definition is found to be inadequate and will be disregarded for the purposes of this paper. 

In this section, we will first look at some recent trends in prosthetic devices with the intention of appreciating the technological advancements in the field of robotic prosthetics. The intention is to cover some examples of non-mechanical robotic prosthetics with neuroprosthesis as the basis, for it's one of the fast-evolving feats of robotic prostheses that are going to assist the future of technology-assisted functional restoration. Topics covered in this section are summarized in table \ref{tab:SectionII}. Starting with a discussion on the recent trends in the field of Neuroprosthetics – This includes a brief study around neural implants and Brain Computer interfaces (BCI). The knowledge acquired through this section will be used for a discussion on control strategies for robotic prosthetic limbs in section \ref{sec:control} of this paper. Some other topics that will be covered in this first section are- advancements in sensory rehabilitation prosthetics, such as visual and auditory prostheses. 


\begin{table}
\scriptsize

  \caption{Summary of the discussion on Neuroprosthetics and their scope, as covered in the section \ref{sec:prosthetics} (Prosthetics).}
  \label{tab:SectionII}
  \begin{tabular}{p{0.05\linewidth} p{0.15\linewidth} p{0.45\linewidth} p{0.12\linewidth}}
    \toprule
    \textbf{S.N. } & \textbf{Keyword} & \textbf{Highlights} & \textbf{References}\\
    \midrule
    1. & Neuroprosthetics \red{(Sec. \ref{sec:neuroprosthetics})} & \begin{itemize}
    \item Interfacing with neurons.
    \item Possibilities offered by the electrical interface between nerves and computers.
    \item Directions of the state of the art in neural implants and Brain-Computer Interfaces.
\end{itemize}   & \cite{ProsBrain}, \cite{FESArm}, \cite{BCIApp}, \cite{BION}, \cite{Ch16ProsBrain}, \cite{NeuralDecoding}, \cite{TouchNP}, \cite{BrainImg}   \\

2. & Visual Prosthesis \red{(Sec. \ref{sec:visualprosthetics})} & \begin{itemize}
    \item Non-mechanical examples of prosthetics.
    \item Sensory rehabilitation through the power of technological advancements in the field of neuroscience and information technology \red{(i.e., subsection of neuroprosthetics)}.
\end{itemize}  & \cite{AgeSightDegen}, \cite{OccuDis}, \cite{OssicReconstr}, \cite{VisProsAssDev}, \cite{VisualProsthetics}\\

3. & Auditory Prosthesis \red{(Sec. \ref{sec:auditoryprosthesis})} &  \begin{itemize}
    \item Non-mechanical examples of prosthetics.
    \item Sensory rehabilitation through the power of technological advancements in the field of neuroscience and information technology \red{i.e., subsection of neuroprosthetics}.
\end{itemize} & \cite{HearingAid}, \cite{Audition}, \cite{CocImpl}, \cite{CochNuclImpl}, \cite{AudPath}, \cite{AudPathWeb}\\


  \bottomrule
\end{tabular}
\end{table}

\subsection{Neuroprosthetics: }
\label{sec:neuroprosthetics}
Neuroprosthetics evolved from the intersection of neuroscience and information technology. \cite{SSPNeuro} These are artificial systems that can control or replace nervous system functions. Neuroscience revealed that the nervous system communicates using electrical signals, and Information theory assisted in the development of tools such as software and control equipment required for capturing and processing these signals.\\



Being the control system for all other organs, issues with the nervous system such as Cerebral palsy, Amyotrophic lateral sclerosis(ALS), Multiple sclerosis(MS), Parkinson's disease, Spinal Cord Injury (SCI), etc. are a major cause of many physical disabilities. The exact problems can vary from issues in the brain to damage to motor nerves such that signals sent from the brain or spinal cord are not properly delivered to the target site in case of motor defects, and damage to sensory organs, sensory nerves, or to the brain or spinal cord itself in case of sensory defects. Neuroprosthetics is the domain of equipment that can potentially tap into the bus of neural signals and manipulate the signals at an organ, nerve, brain, or spinal cord level. Popular terminology from this domain includes Neural Implants, Brain Computer Interface, and Electroencephalography, etc. Research and Development in this area has shown that neuroprosthetics can help fix a number of sensory and motor disabilities including blindness \cite{VisProsAssDev}, deafness \cite{CocImpl}, partial paralysis \cite{ProsBrain} \cite{BION} \cite{FESArm}, Numbness \cite{ Ch9NPLeg}, Muteness \cite{ Brain2Speech} etc. 

\red{Unlike other organ systems of our body, for defects/damages to nervous system, medical healing (regeneration) and transplantation are not viable options.} Regenerative medicine is capable of replacing lost tissue or an organ with artificially regenerated, but otherwise identical tissue, and research has successfully shown such techniques in wound healing and orthopedic applications. There is active research focused on regenerating full organs such as heart, kidney, liver, etc. in-situ as well as ex-situ followed by transplantation. When it comes to neurons, their connections with one another are just as crucial as the neurons themselves.
. Neural tissue also has a very low intrinsic regeneration capability. Unlike a liver or kidney transplantation, it is generally considered farcical to do a spine or brain transplant \cite{Ch16ProsBrain}. This further emphasizes the relevance of neuroprosthetics.


On one hand, neural prosthetics can be used to interpret electrical neural activity from the brain to artificially generate control signals for controlling a paralyzed limb, controlling a prosthetic arm, or moving a computer cursor, etc. Likewise, On the other hand, neural prosthetics can be used to restore sensation, for example in an amputee wearing a prosthetic limb external sensors are used to capture physical stimulus from a desensitized body part to then relay corresponding electrical stimulus directly to nerves. Such devices can potentially mimic the sensation of touch, pressure, temperature, etc.

Capturing and processing of brain signals is called Brain Computer Interface (BCI). The potential of BCI was demonstrated as early as 2002 wherein, employing brain-computer interfaces, monkeys were trained to manipulate robotic limbs and a computer cursor to different target locations \cite{MonkeyCursor}. For BCI, typically external (non-invasive) sensors are used when the requirement is to capture neural signals reflecting the activity of many neurons (low-resolution applications). Invasive sensors are surgically implanted arrays of electrodes that can use waveform shape differences to discriminate individual neurons. The pattern of neural activities is observed before and during a movement. A map is obtained between actual limb movements and associated neural activity by decoding the collected pattern using various computational algorithms. This map can be used to generate appropriate control signals artificially for driving a paralyzed or prosthetic arm. This will be an example of motor prosthesis application of neuroprosthesis. Likewise, the control signal generated may be linked to positioning a computer cursor to a specific position on the screen such as for operating a virtual keyboard – this would be labeled as a communication prosthesis \cite{NeuralDecoding}. 

\begin{figure}

     \centering
     \begin{subfigure}[b]{0.5\linewidth}
         \centering
         \includegraphics[width =\linewidth]{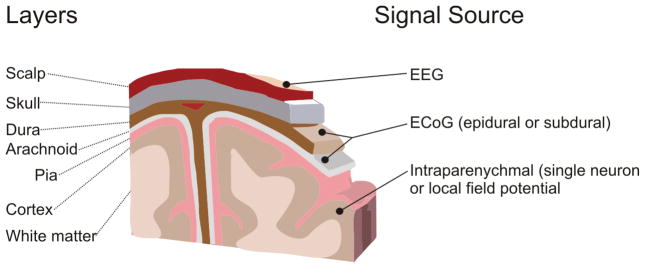}
    \caption{Types of BCI based on site of interfacing.\cite{EEGEvolution}}
    \label{fig:BCIEvolution}
     \end{subfigure}
     \hfill
     \begin{subfigure}[b]{0.4\linewidth}
         \centering
         \includegraphics[width = \linewidth]{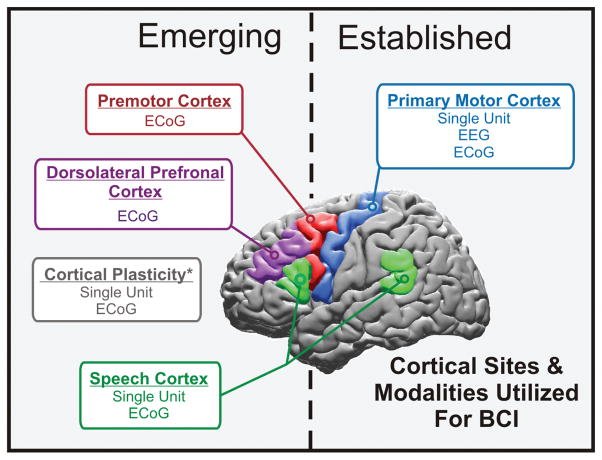}
     \caption{ Typical target sites of brain for BCI.\cite{EEGEvolution}}
     \label{fig:BCISites}
     \end{subfigure}
     
        \caption{  Brain Computer Interface - Strategies }
        \label{fig:bci}
        \Description[Brain Computer Interface - Strategies ]{This figure shows the different types of BCI based on the sites of implantation. Thesubfigure (a) shows these types based on the depth of implantation in the brain. The
subfigure (b) shows the types based on targetted interface with various sections of the brain.}

\end{figure}

For clinical applications, currently, there are three types of BCI platforms, categorized primarily on the basis of the site of the interface (Fig. \ref{fig:BCIEvolution}). These categories are noted as Electroencephalography or EEG (Brain signals collected from the scalp), Electrocorticography or ECoG (Signals collected from the epidural or subdural region), and Intraparenchymal (signal from local field potential or from single neuron). EEG is the most popular mode of BCI under study due to the convenient, safe, and inexpensive nature of the technique. The signals obtained through EEG has very poor spatial resolution as it consists of resultant signal from a large number of neurons. Since the amplitude of these EEG signals is very small, this signal is also prone to noise from other sources such as electromyography signals outside the brain, etc. Despite the limitations, Many relevant studies have been conducted that show promising applications in clinical as well as outside settings. Studies have shown success in using EEG for controlling computer cursor accurately in 2D \cite{2DBrainMouse1}, \cite{2DBrainMouse} and 3D \cite{3DBrainMouse}. However, accurate control requires substantial training (around 50 sessions of a 30-minute training routine for control in 2D \cite{ EEGEvolution}.)

In ECoG electrodes have to be surgically implanted into the region and the obtained signals have up to 5 times larger spatial and temporal resolution (0.125 cm compared to 3 cm in EEG, and up to 500Hz compared to 0-40 Hz in EEG) \cite{EEGEvolution}. EEG has allowed for studies of mu(8-12 Hz) and beta (19- 26 Hz) brain waves, Whereas ECoG has enabled close study of Gamma brain waves (25-80 Hz) that are associated with many aspects of motor and speech functions in humans. NeuraPace is an innovative device based on ECoG BCI which aims at the identification and abortion of seizures \cite{Neurapace}.

Single neuron-based systems aim at implanting an array of minute electrodes into the cortical layers at around 1.5 to 3 mm. This allows for recording signals from individual neurons. The size of these electrodes is approximately 20 micrometers in diameter and these are inserted into brain parenchyma where extracellular action potential in the range of 300 microvolts can be measured from neurons around 10 to 100 micrometers away from them. These implants have been successfully tested on monkeys \cite{SingleNIMonkey} and humans \cite{SingleNIHuman1}, \cite{SingleNIHuman2}.  Despite the promising results, there are two big problems with such implants. One is the risk of CNS infections along with the unavoidable vascular local neural damage, the Other is the reactive cell responses exhibited by the movement of astrocytes and microglial cells toward implantation cite \cite{SingleNIIssues}. Devices such as Neuralink \cite{Neuralink} are adding new dimensions to the possibilities that BCI is capable of.

One problem with these brain implants is that their embedding requires opening the skull and placing the implant directly into the brain. This is a complicated procedure that is new to most of the neurosurgeons. These brain implants are also known to scar the tissue around them which results in reduced sensitivity over time. A different approach to brain implants was recently approved by the US FDA for initial clinical trials. This device, called Stentrode\texttrademark (developed by Synchron) \cite{stentrodeog}, \cite{stentrodefull} is in the form of an endovascular stent that is inserted into the body of the patient through the jugular vein and snaked through the vein to land into the superior sagittal sinus which is large between the two brain hemispheres. The stent is installed at a site near the primary motor cortex, which is of focus for most BCI studies. Once the stent is located correctly, it blooms open and gets attached to the walls of the blood vessel with a thin wire routing through the blood vessel to a controller module implanted into the patient's chest. This controller module is responsible for relaying the signals to an external receiver. Application of such a system has been successfully shown in a fully paralyzed ALS patient who is able to communicate by moving a computer cursor using his thoughts. 
The installation of an endovascular stent is a usual practice for most cardiovascular and neurosurgeons, this eliminates the requirement of specialized training for a neurosurgeon for embedding the brain implant. An endovascular implant also improves in performance over time as the stent heals into the walls of the blood vessel.  

Most of the brain signals currently used for BCI are collected from the primary motor cortex because the physiology of this region is most clearly linked to the user's intent of limb movement and interaction with the world. While BCI targeted at the primary motor cortex has seen many promising studies for patients of SCI and neuromuscular disorders, Other parts of the brain (Fig. \ref{fig:BCISites}) such as the premotor cortex, dorsal prefrontal cortex, speech cortex, etc. are now being considered for BCI and are expected to reveal ways to aid patients of other neural conditions. Although these other areas can be studied only via invasive methods only \cite{EEGEvolution}.

Here communication prostheses are an interesting side track from this study – Communication prostheses are concerned with providing means to fast and accurate communication which may not necessarily be in the form of natural voice or typing. People with a total disability such as ALS and patients with the "locked-in" stage can communicate using their eye movements (which are generally not affected by the disease). This can be achieved by – say, using eye tracking to control an onscreen keyboard by moving a computer cursor to type out words.

The field of neuroprosthetics has been receiving a lot of research attention recently and the increased commercial interest especially in the applications related to restoring peripheral sensory functions including vision (Section \ref{sec:visualprosthetics}), hearing (Section \ref{sec:auditoryprosthesis}), and touch (Section \ref{sec:eskin}) etc. is aiding in the development of an even broader research base for technological developments. \cite{ProsBrain} Some of these applications are covered in the following paragraphs.

\subsubsection{\textbf{Visual Prostheses:}}
\label{sec:visualprosthetics}
Prostheses that Aid/Enable vision. While simpler items such as intraocular lenses might be classified as visual prostheses, The term Visual prostheses is typically used to define devices that use artificially generated electrical impulses to generate vision-compatible stimuli along the signal pathway of vision to restore lost visual function. Numerous groups have demonstrated, that patterned electrical stimulation along the vision pathway can evoke patterned light perceptions.\cite{patternedsight}, \cite{patternedsight1}, \cite{patternedsight2} This constitutes the foundation of visual prostheses which are aimed at restoring vision in blind people. The approaches in visual prostheses vary according to the site chosen for electrical stimulus based on which, current studies can be classified into epiretinal, subretinal, choroidal, and cortical visual prostheses.

Most of the commercial interest in visual prostheses is targeted towards retinal applications \cite{AgeSightDegen}. Non-retinal locations in the visual pathway such as the visual cortex and lateral geniculate nucleus are still under experimental stages. Currently, there are over a couple of dozen clinical trials for visual prostheses registered in the United States National Library of Medicine \cite{AgeSightDegen}. A noteworthy work from the commercial area is Argus II Retinal Prosthesis System (Argus II) by a company called{Second Sight}, which is the first FDA-approved (2013) commercial offering for epiretinal visual prosthesis \cite{argus2fda}. Argus II captures the video feed through a small camera fitted on a pair of glasses, and the processed signals are wirelessly transmitted to a retinal prosthesis with 60 implanted electrodes. The electrodes convey the transduced electrical impulses onto the epi-retinal area, from where the signals are sent through the optic nerve to the brain. Thereby bypassing the degenerated photoreceptors of the retina \cite{argus2details}.

The number of diseases a visual prosthetic can cure is correlated with how central (i.e., closer to the brain) the location of a prosthetic is \cite{OccuDis}. Inherited blindness is predominantly caused by a disease called retinitis pigmentosa. Over 1.5 million people are affected by this disease worldwide (1 per 4000 births) \cite{Ratina5}. However, retinal prostheses can not treat glaucomatous blindness, as in this case, the damage is to the optic nerve. 

Each approach to visual prostheses has its own set of merits and demerits. Any intraocular implants are likely to cause chronic eye inflammation however the brain is fairly inert to long-term implantation of foreign material \cite{VisProsAssDev}, this represents the major problem with retinal prostheses.  Any transcranial application is more complicated to operate, which adds difficulty in choroidal and cortical visual prostheses. Visual prosthesis approaches around the optic nerve have been studied for placing a cuff electrode array in the ocular orbit, but the optic nerve is enclosed within all three meningeal sheaths, requiring a stronger electrical potential for stimulation by the cuff electrode \cite{VisProsAssDev}.
Research in the field of visual prostheses is continuing at a very fast pace. From the demonstrations of initial proof of concept around 1996 to the availability of the first epi-retinal commercial offering in 2013, Within a decade and a half \cite{Humayun2006}. The field of visual prostheses is driven by active research to commercial availability, this itself is inspiring. 


\subsubsection{\textbf{Auditory prostheses:}}
\label{sec:auditoryprosthesis}
Like visual prostheses, Auditory prostheses are also of various kinds depending on their area of application. The most basic auditory prostheses function as a voice amplifier, which can be tuned to the specific frequencies that the person has a hearing deficit. Further prostheses that deal with hearing disability at the middle ear and cochlear (inner ear) level are also studied. Middle ear prostheses target mechanical compensation for lack of mobility/absence/necrosis in the middle ear bones- Incus, Stapes, and Malleus and sometimes complete fixation of the ossicular chain might be required \cite{OssicReconstr}.


       
     

For a long time, it was deemed impossible to target hearing disabilities beyond the scope of mechanical fixation in the middle ear. The inner ear is composed of a complex anatomical structure called the cochlea. The cochlea is a spiral-shaped hollow ossicular structure containing hair cells and approximately 30,000 nerve fibers \cite{Audition} that transduce the mechanical sound waves into electrical impulses for conduction to the brain, along with being responsible for balancing in humans. It appeared to be daunting to stimulate such a complex instrument with a few artificial electrodes to restore hearing. Eventually, it was discovered that not all parts of hearing are related to the complicated physiological response of the cochlea. 

Cochlear hearing impairment can be treated with cochlear implants. These systems contain a microphone, processor, and transmitter in a unit placed behind the ear along with an implanted unit in the inner ear. The implanted unit consists of a receiver which is typically implanted just below the skin, and electrodes implanted in the cochlea \cite{CocImpl}. Electrodes are inserted into a region of cochlea called the Scala tympani. Patients suffering from age-related hearing disabilities are good candidates for the cochlear implant because direct stimulation of the auditory nerve by such implants bypasses the cochlear hair cells, which are affected by age-related hearing loss. Nowadays, systems combining traditional sound amplification-based hearing aids with cochlear implants are commercially available. These systems typically target the low-frequency hearing compensation through the traditional method, and the high-frequency regions are targeted by direct cochlear stimulation using electrode arrays.

Current-day research targets hearing loss at even deeper levels. Auditory Brain-stem Implants (AIBs) exist \cite{AudPath}.  For some patients, hearing loss is the result of the destruction of the auditory nerve either through surgery or trauma, which renders cochlear implants of no use in restoring hearing. AIBs have arrays of electrodes that can directly stimulate the cochlear nuclei (in the brain) bypassing the auditory nerve \cite{CochNuclImpl}.

 \subsection{ Scope of this review }

\red{In the previous section, neuroprosthetics are in some depth to form the grounds for a better discussion on sensory rehabilitation and control strategies through sections \ref{sec:control} and \ref{sec:paralleltrends}}. Owing to the vastness of the term robotic prostheses, the further review of robotic prosthetics is focused on the mechanical motion\red{, control strategies, and advanced sensory integrations for robotic prosthetics.} To add, the discussion will be confined primarily around \emph{Robotic Prosthetic Limbs}.  We regularly see people in our society who go through their day-to-day lives using a hand or a leg that’s made up of plastic and metal. Sure, such a passive prosthetic limb does complete a person’s stature, which helps them be more comfortable in society. However, these prosthetic limbs have more to be desired. Current-day robotics has the potential to give a person a functional limb made of plastic, metal, motors, and wires.  The further discussion includes a new trend in soft robotics-based actuators for prosthetic applications, and control strategies for these robotic prosthetic limbs, wherein this paper also expands on the applications of neuroprosthesis in this field.
This research revolves around these two pointers and the study in these fields is concluded in sections \ref{sec:motor}, \ref{sec:control}. Later, in section \ref{sec:paralleltrends}, A study on some of the parallel trends in active research in this field will be covered.


\section{Motor systems for Robotic Prosthetic Limbs}
\label{sec:motor}
Conventionally, motion in a robot is associated with a motor, and it's generally hard to think of a feasible alternative. \red{ Hence,  whenever} we think about a robotic prosthetic limb, typically an assemblage of motors, gears, and circuitry comes to mind. \red{Recently}, we have grown to classify the field of robotic devices into two broad categories: Hard robotics and Soft robotics,  \red{where the latter is a highly active field of research with many potential applications}. Soft robots are primarily made of soft materials, components, and structures. Compared to robots composed of hard components, they are better able to safely interact with and adapt to their immediate environment. Soft robots are often described as bio-inspired robots as they generally take cues from, and base their functioning on, biological structures. Because of this, they provide the best answers to needs related to soft touches, secure interactions with humans, holding and handling small objects, etc. In the following sections, The need to explore soft robotic actuators for robotic prosthetic applications will be understood first. Then a bit deeper dive into the domain of soft robotic actuators and an analysis of the feasibility of various options within this domain is done.\\

\subsection{\red{\textbf{Current state of robotic prosthetics:}}}
\textbf{Hard robotics} is more or less the representative of the conventional form of robotics wherein motion is generated at the point of actuation through rigid electromechanical motors or chemo-mechanical engines and the power is transmitted through a constrained network of rigid linkages. These kinds of robots are solid and hard, thereby are capable of huge amounts of destruction upon any malfunction or unregulated operation. These robots are often driven by electrical motors through appropriate gearbox arrangements. It's a generally agreed-upon fact that the energy efficiency of such systems is not optimal. The stall current, backlash, etc. contribute to this inefficiency, and in a portable system such as a robotic prosthetic limb, a more efficient mechanism for motion generation is desirable. While harmonic drives and high-performance servo motors are considerable options, research is being focused more on the non-convectional soft robotics-based options in motion generation. \\



A brief account of the reasons associated with the recent popularity of soft robotics is given below:
\begin{itemize}
    \item Inorganic association of a hard machine part with a human is undesirable and may lead to harmful consequences. Hard robotics-based prosthetic limbs can also  not be given powerful motors/gears, This has multifold reasons behind it: 
    \begin{itemize}
    
       \item Unlike our muscle system, the forces on such motor-driven limbs are more concentrated. They also have high inertia \cite{MRFeasibility}. Since the device has to serve as an extension of its human, this limits the amount of force that one can implement through such robotic prosthetic limits.                  
       \item More power also requires heavier components such as bigger motors, heavier construction, and more powerful batteries. This limits the practicality of such an initiative even further.
        \item Powerful hard robotics-based prosthetic limbs that lack any mechanical compliance can be a safety hazard also. A glitch in the motor controller may lead to lethal consequences for the person using the limb, as well as for the people nearby. \cite{BiomedSoftRob}   
        
         \item Social and aesthetic issues associated with such an inorganic prosthetic limb also make them less desirable. Non-mechanical cosmetic prosthetic limbs are often preferred over the more functional robotic prosthetics as the former can be crafted to match a person's appearance elegantly, which makes the acceptance process more fluid. \cite{Aesthetic}
                  
    \end{itemize} 
     \item It is also noted that typical robotic prosthetic limbs are suited for replacing the extremes of limbs only, such as for trans-radial or trans-tibial amputation. This is for 2 reasons:
    \begin{itemize}
             \item Reaction forces concentrated at the root become excessive and are potentially detrimental to the user.
        \item Most of these commercially available robotic prosthetic limbs generally rely on sensors around the residual muscles of amputees (Described in better detail in the section \ref{sec:control}: Control Strategies). This makes the robotic limbs' motion more predictable and easier to control from an engineering perspective. However, It relies on the person being able to produce muscular contractions that were otherwise obsolete after amputation- and this may require serious psycho-physical therapy before a person is capable of making such actuation properly. Anyhow, this is a shortfall of the control strategy rather than hard robotics, so this issue will be considered in section \ref{sec:control} of this paper.
      
    \end{itemize}
\end{itemize}
\red {There are several other reasons, not related to hard robotics specifically, that are limiting the acceptance of robotic prosthetics in general among potential beneficiaries. One of the most prominent reasons is expensive Research and Development associated with the person-to-person customization which is required for prosthetic limbs is another problematic area. Recent advancements in 3D scanning and 3D Printing have helped a bit in this area. Maintenance of such a robotic part which is driven by conventional motors and gears is also not ideal. Another important fact is \emph{people grow, things don't.} Prosthetic limbs are required to be reshaped and remolded as the person grows. For young kids, this can be required at intervals as short as 8-10 months. Typical robotic prosthetic limbs are more complex and expensive. Therefore, replacing them at such frequency makes them much more inaccessible, especially for the growing kids. These are some of the concerns that are to be fixed for large-scale acceptance of robotic prosthetics among amputees. }\\

\begin{table}[H]
\scriptsize 
\caption{Comparative analysis between hard and soft robotic actuators for prosthetic applications }
\label{tab:hardvssoft}
\begin{tabular}{p{0.05\linewidth} p{0.45\linewidth} p{0.45\linewidth} }
\toprule 
\textbf{  } & \textbf{Hard Robotics} & \textbf{Soft Robotics} \\
\midrule 
\textbf{Pros} & \begin{itemize}
    \item Relatively generic, Proven, Intuitive, and reliable technology.
    \item Available commercial options and active commercial interests. 
    \item Potentially cheaper than the counterpart.

\end{itemize} & \begin{itemize}
    \item Wide range of technologies to choose from for specific needs.
    \item Organic and bio-inspired designs.
\item Huge scope for research and development from actuator design, material science, etc.
\item Characteristically soft and compliant.
\item Lighter and more energy-efficient options are available.
\end{itemize} \\


\textbf{Cons} & \begin{itemize}
    \item Inorganic mechanical design.
\item High force concentration.
\item Limited scope for research and further development.
\item Rigid and non-compliant.
\item Heavier and have higher maintenance requirements.
\item Mechanical nature may have aesthetic and social implications.

\end{itemize}

& \begin{itemize}
    \item Engineering may be complicated and often actuation principle is based on a molecular level.
    \item Most options are still under research and produce a lower amount of force.
    \item Evolution of the field is based on contributions from multiple genres of science and technology.
    \item The Current state of technology has scope for a lot of research and development.
    \item No, or very limited commercial attention
\end{itemize} \\
\bottomrule
\end{tabular}

\end{table}

\subsection{\textbf{\red{Recent Advancements in Soft Robotic Actuators}}}
The field of \textbf{Soft Robotics} brings interesting opportunities to address some of the inherent flaws associated with their Hard counterpart. \red{A summarized comparison of Hard and soft robotics is given in table \ref{tab:hardvssoft}.} Soft Robotics is a huge field of structures, sensors, and actuators that are based on a diverse array of scientific principles but share one property in common, they are all physically soft and compliant.  Here will discuss the major technologies behind soft robotic actuators. These technologies have a lot of potential and active research is underway in engineering, material science, and fabrication technologies to utilize these actuators for powering various kinds of prostheses in the future.

\begin{enumerate}

    \item \textbf{Flexible Fluidic Actuators: }
    
    Flexible Fluidic Actuators are compliant contraptions made out of soft materials that change in shape, viz. elongation or contraction, upon being provided with a forced fluid (Gas or liquid). Superficially, they may appear as functional analogs of animal muscles. The first cite of such actuators originated way back in the 1950s with McKibben Muscles. With time, the simple technology grew through multiple iterations such as Rubbertuator Arm (Bridgestone Inc.) to really complex and life-like implementations such as in Shadow C3 Dexterous air muscle hand \cite{shadowhand}. 
    
    There are a lot of types of FFAs based on their construction, however, conceptually they are mostly the same, i.e., \textit{pressure exerted by the fluid is used to create motion which is directed using a suitable compliant mechanism.} The functioning principle behind the most common FFA models is as follows: When the tube is pressurized by the inwards flow of a fluid (which can be a variety of kinds of liquids or gases), it expands in volume. The sleeve is driven by an axo-radial compensatory deformation due to the nature of the sleeve. Common FFA designs can be classified into three categories, as depicted in figure [\ref{fig:ffa}].
    
      \begin{figure}[H]
             \includegraphics[width = \linewidth]{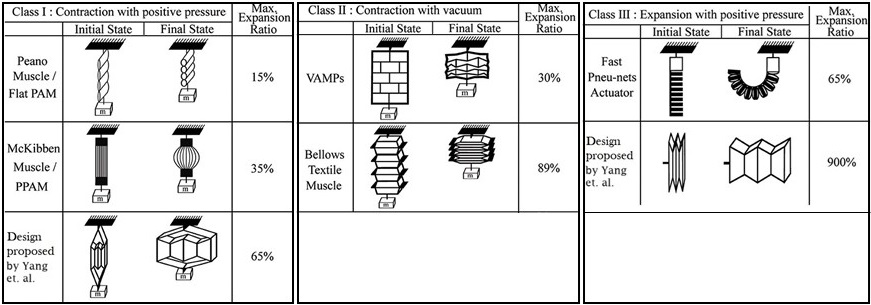}
             \caption{\red{\textbf{Fluid Filled Actuators: } Visual illustration of classification of FFAs by Yang et al.} \cite{FFATyp}}
             \label{fig:ffa}
            \Description[Fluid Filled Actuators]{ This figure illustrates some example designs of Fluid Filled Actuators as published by Yang. et al.}             
        \end{figure}

    These muscles may be driven pneumatically or hydraulically, With the former facilitating a more compliant system and the latter being more rigid and predictable. In recent years, Fluid Filled Actuators (FFA) has been used vastly in the domain of surgical robots due to the unmatched centricity that these offer compared to the conventional robots \cite{BiomedSoftRob}. These muscles can be arranged in sophisticated patterns and can be coupled with various restraining methods to produce life-like muscular behavior. A nice take on origami-inspired FFAs was covered in \cite{FluidOrigami}. 
    
    Despite their many advantages, FFAs have typically had some fundamental problems associated with their operation. FFAs require external pumps/ Compressors to function. This limits the applicability and practicality of using such muscles for the fabrication of a prosthetic limb. Another problem is that the more pressurized the actuator is, the more rigid it’ll get, this may be a cause for concern as the system has to work in proximity to a human being. The third problem is the failure associated with the wear and tear of these muscles, as their functioning is fundamentally dependent on the system being leak-proof. 

    Recent advancements have attempted to address these problems. Thin McKibben Muscles (TMMs) \cite{TMMOrig} have an inner diameter of the order of 1 mm and are highly flexible even upon actuation with pressurized fluids. Their thin size allows bundling them into stronger units. This solves the problem of pressure stiffening. ElectroHydroDynamics-based solid state pumps have been fabricated recently \cite{StrPump}. These tiny pumps are designed to operate single strands of TMMs individually. One pump along with a TMM filament was shown to weigh 2g and was capable of delivering a maximum stroke of 4 mm and blocked force of 0.84 N \cite{ElectricalTMM}. The relatively high voltage requirement for the operation of these EHD-based pumps is not ideal, and the complicated fabrication of such EHD-based pumps demands technological advancements prior to any kind of scaled production\cite{ElectricalTMM}.

    The problem of loss of functionality due to wear and tear is addressed by the introduction of self-healing properties to the elastomer or to the fluid used. Self-healing FFAs which are capable of healing macroscopic damages to the elastomer by application of heat are published by Terryn et al. \cite{SelfHealReview}. A more recent take on hydrogel-based FFA that can heal themselves when damaged, is given in \cite{HydrogelPMM}. Various self-healing polymers for soft robotic applications are covered in  \cite{SelfHeal}.\\

    %
          

    \begin{figure}[H]
     \centering
       \begin{subfigure}[b]{0.45\textwidth}
             \centering
             \includegraphics[width =\linewidth]{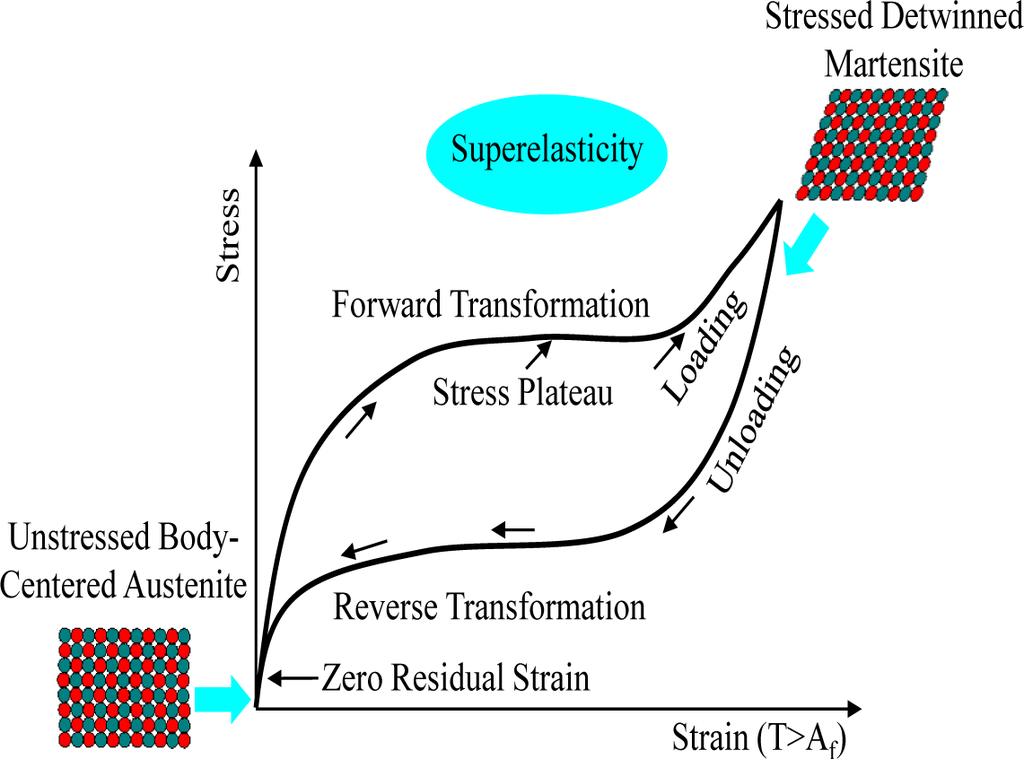}
             \abovecaptionskip=0pt
             \caption {Nitinol Characteristics\cite{StressStrainSMA}}
             \label{fig:SMAStressStrain}
          
             \end{subfigure}
     \begin{subfigure}[b]{0.45\textwidth}
         \centering
         \includegraphics[width=\linewidth]{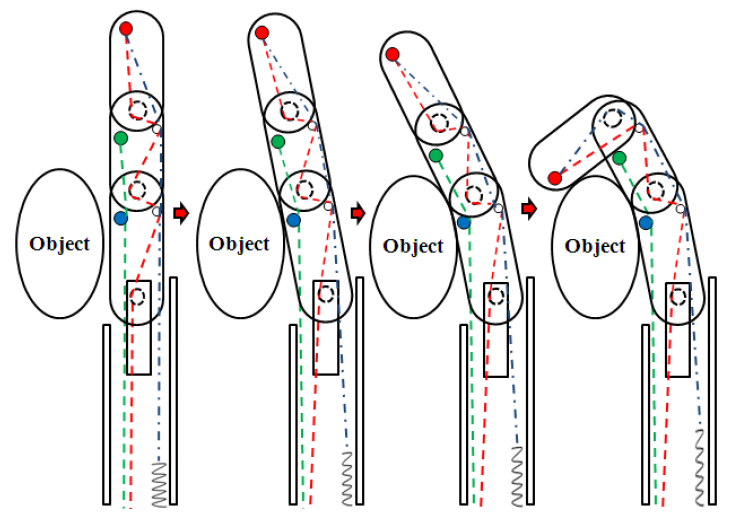}
        \caption{Compliant Grip}
        \label{fig:CompliantGrrip}
     \end{subfigure}
        \caption{ (a) Stress vs strain curve of SMA. \red{$A_f$ temperature (Or Austinite finish temperature) represents the temperature at which the material is converted out from the austenite phase.} (b) SMA Engineering Designs: SMA-based prosthetic finger actuation schema as presented in \cite{SMAArm}.\red{The extension of fingers is based on a spring situated in the backside of the palm. While individual links are contracted by activating the respective sma wires (blue: base link, green: middle link, red: fingertips)}}
        \label{fig:SMAHand}
        \Description[Stress Strain Curve of SMA and an SMA based prosthetic hand]{The subfigure (a) illustrates Stress vs. strain curve of SMA. Af temperature (Or Austinite finish temperature) represents the temperature at which the material is converted out from the austenite phase. The subfigure (b) illustrates SMA engineering design of SMA-based prosthetic finger actuation schema as presented in reference [3]. The working of the illustrated design is explained as follows: The extension of fingers is based on a spring situated in the backside of the palm. While individual links are contracted by activating the respective SMA wires (blue: base link, green: middle link, red: fingertips).}
    \end{figure}

    \item \textbf{Shape Memory Alloy/Polymer: }  
     
     Shape Memory Alloys are metallic alloys that can revert to a pre-configured shape by heating after a plastic deformation. The reason for this is associated with the particular mechanical characteristics connected to the various stable crystalline phases of the material. They have a critical transition temperature at which they are formed into a specific shape. After plastic deformation, if the material is brought back to the transition temperature, they are capable of getting autonomously reformed into its initial geometric form. These materials are also studied for their super elastic nature around the transition temperature \cite{BiomedSoftRob}. A typical example of SMA is Nitinol (Ni14Ti11) \cite{nitinolog}.
       
    Another option in smart/shape memory materials is Shape Memory Polymers (SMPs) which are nonmetallic, organic polymers that function similarly to SMAs but are easier to produce and represent a more compliant option for biomedical applications. Some examples of Shape Memory Polymers include polylactic acid (PLA), polytetrafluoroethylene (PTFE), ethylene-vinyl acetate (EVA), etc. A great resource for studying the field of Shape Memory Polymers is the book \cite{SMABook} wherein SMPs are covered in terms of principle, properties, and fabrication. 
     
     SMAs are capable of producing strong actuation forces. They are relatively denser compared to EAPs (at 5-6 gm/cc compared to 1-2 gm/cc for EAPs) but the force generation is exponentially better - ~700 MPa compared to 0.1 - 3 MPa in EAPs \cite{NASAEAP}. The downside of SMAs compared to EAPs is lower strain (~8 percent) and slower response times (typically in seconds, compared to micro to milliseconds in EAPs). SMAs can be composed into Smart Composites (Shape Memory Composites (SMCs)) of suitable sort, using compliant mechanisms and clever mechanical engineering to compensate for the downsides.
     
     At lower temperatures, SMA's exist in \textit{Martensite Phase,} where they are compliant and can be deformed. At higher temperatures, SMAs get into the \textit{Austenite phase,} where they exhibit their shape memory characteristic and their super elasticity. \red{The phase transition temperatures for shape memory alloys are very crucial for their applications. These temperatures are an intrinsic property of the material which include austenite start ($A_s$), austenite finish ($A_f$), martensite start ($M_s$), and martensite finish ($M_f$) temperatures. These four transition temperatures of a popular NiTi-based shape memory alloy, Flexinol are 52.54, 60.90, 44.78, and 32.94 degrees Celsius respectively \cite {flexinoltemps}.  Nitinol alloys with lower $A_f$ (around 11 degree Celsius) is used in intravascular stents.\cite{stentnitinol} } An inspiring bio-mimicking SMA-based Artificial Bicep Muscle Composition as covered in \cite{SMAElbow}. \\

     A characteristic Stress vs. strain curve of Nitinol SMA is given in figure [\ref{fig:SMAStressStrain}]. The curve illustrates an interesting feature, i.e., Super-elasticity, which is also a prominent field of research for Shape Memory Alloys. \red{The super-elastic effect are principally different and the fact that SMAs can exhibit the super-elastic effect is due to the transitioning of the material from load-induced austenite phase back to martensite phase upon unloading, given temperature is above $M_f$ (martensite finish).\cite{supelastvssma}} SMAs do not provide very fast actuation, but due to their lightweight nature and modularity their use is gaining popularity in engineering applications including prosthetic limbs (Fig. \ref{fig:CompliantGrrip})
     
    \item \textbf{Electroactive Polymer: }
    
     Electroactive polymers (EAPs) are Soft materials and structures that exhibit a deformation under an electric field. Electroactive Polymers (EAP) are considered good artificial options that emulate the functioning of actual biological muscles. Being polymers, they possess low density, mechanical flexibility, ease of processing, and mass production capabilities \cite{OrigEAP}. Besides exhibiting a change in shape upon electrical actuation, they can also sense mechanical strain, Which can enable homologous implementation of these artificial muscles to their natural counterparts. Their potential in emulating human muscles is widely anticipated in the scientific community. This has resulted in the establishment of Worldwide EAP(WW-EAP) Web hub \cite{WW-EAP}.

    Based on the activation mechanism, EAP actuators are divided into two major groups- (a) Ionic EAP and (b) Field-activated EAP. The key differences are summarized in table \ref{tab:EAPType} and a detailed discussion of these categories and their subtypes is given below.

        \begin{enumerate}
                 \item \textbf{Ionic EAP:} These actuate based on the diffusion of ions (Fig. \ref{fig:IPMC}). Their general structure is constructed with \emph{an electrolyte film, which is sandwiched between two flexible electrodes.} They exhibit deformation in the form of bending which is due to accumulation of the ions closer to one electrode, causing that side to bulk up, and the other to shrink.  They can be made to bend the other way just by reversing the polarity across electrodes. They have low activation potential (1-2 volts) and generate large displacements in the form of bending. They require constant maintenance of electrolytes, which is a major disadvantage. Traditionally, they have been susceptible to electrolytic degradation and require constant maintenance of electrolytes. However, recent advancements such as solid polymer electrolytes, bulky gel ionomers, etc. have successfully addressed these flaws.\cite{TempAssSACP} \cite{IonicEAPFab}
                
                
               The electrical field causes the ions in the electrolyte to diffuse based on the polarity. This diffusion can take place at very low voltages (requiring only 1–2 V at the lower end). They typically require encapsulation by an aqueous medium for sustained operation, however, currently, Solid Polymer Electrolyte based Ionic EAPs are available\cite{TempAssSACP} \cite{IonicEAPFab}. 
                
               Ionic EAPs are considered to be the ideal choice for microactuators in surgical robots due to their low voltage precision control and limitless dexterity compared to their hard robotics counterparts. When considering Ionic EAPs for biomedical applications, the materials used for the fabrication of EAP become crucial, considering that many published recipes for the fabrication of Ionic EAPs require potentially toxic Ionic Liquids. Nafion-based, ionic liquid-free, bio-compatible Ionic EAP fabrication is being studied extensively.\cite{TempAssSACP}. 

              Most of the Ionic EAPs are tri-layered structures, wherein the 2 external layers are made of conductive materials and the inner layer consists of ions suspended in some matrix (Liquid/Gel/Solid).

                The origin of EAPs can be traced back to the 1900s and based on the general construction many materials have been considered and studied for the electrode (The conductive layers) as well as the ion exchange membrane (Ionomer layer) to enhance the electromechanical performance of these actuators. All varieties of Ionic EAPs are based on the same aforementioned principles. Major variables when it comes to fabrication are Electrode, Ionomer, and Fabrication Methodology \cite{IonicEAPFab}. 
                \begin{figure}[H]
                     \centering
                     \begin{subfigure}[b]{0.45\textwidth}
                         \centering
                          \includegraphics[width =\linewidth]{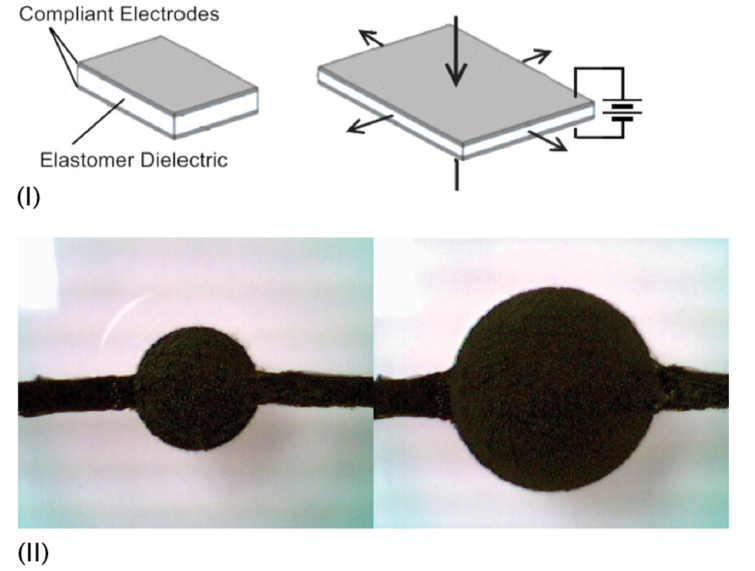}
                         \caption { Dielectric EAP \red{:Arrows indicate axial compression from electrostriction, causing material expansion in the equatorial direction.} \cite{OrigEAP}}
                         \label{fig:DEEAP}
                     \end{subfigure}
                     \hfill
                  \begin{subfigure}[b]{0.45\textwidth}
                         \centering
                         \includegraphics[width=\linewidth]{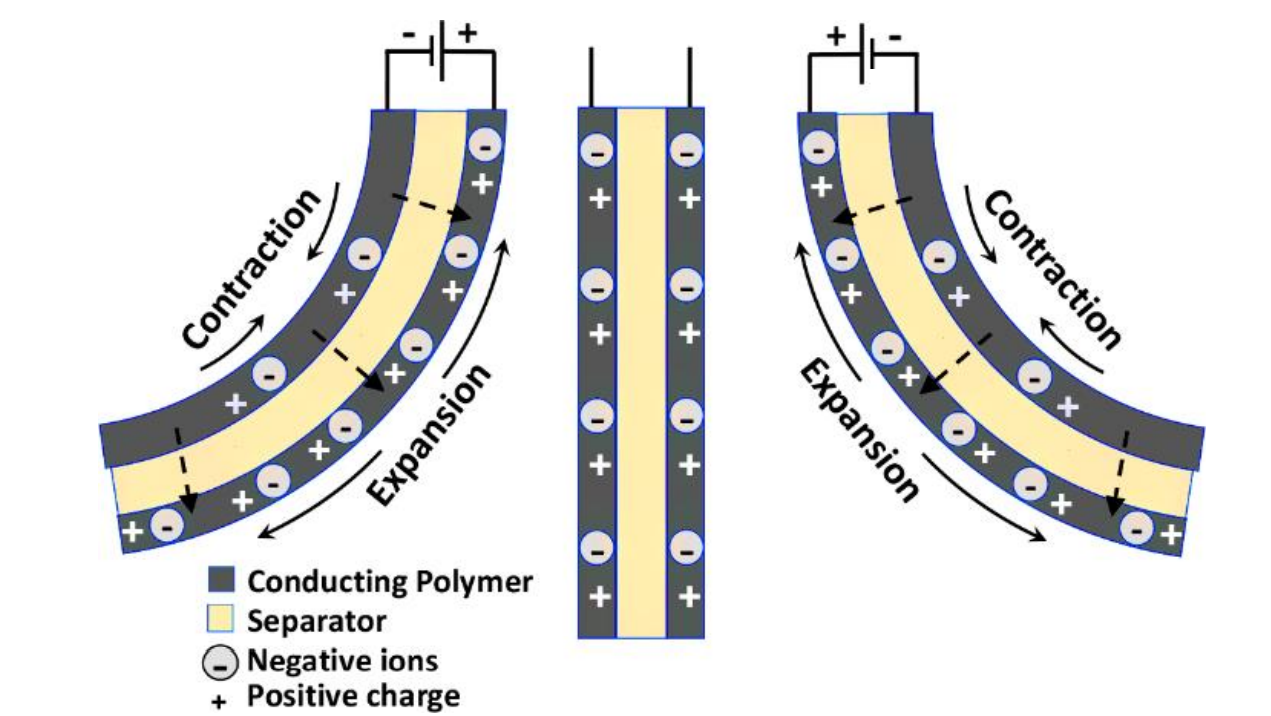}
                        \caption{Ionic EAP \red{: Arrows in the ionomer layer depict ionic movement, causing compression and expansion as shown by peripheral arrows.}\cite{IonicEAPReview}}
                       \label{fig:IPMC}
                  \end{subfigure}
                     \caption{Working of Popular EAPs.}
                       \label{fig:EAPs}
                       \Description[Electroactive Polymers]{This figure illustrates the actuation of the two types of electroactive polymers as described in this paper. In the subfigure (a), actuation of Dielectric EAPs is described where the arrows indicate axial compression from electrostriction, causing hte material to expand equitorilly. The subfigure (b) illustrates th eactuation of Ionic EAPs where the arrows in the ionomer layer depict ionic movement according to the applied polarity, causing the expansion of the side with bulkier charge body and contraction on the opposite side}
                \end{figure}

              \item \textbf{Field Activated EAP:} These kinds of EAPs exhibit their deformation due to the principle of Electrostriction. Their different types vary in construction. Electrostriction is defined as the property of all Dielectric materials (Insulators) due to which they manifest a slight change in shape upon application of Electrical Field. Unlike Ionic EAPs, changing the polarity of the field does not have any effect on the actuation. These are also primarily studied in two subcategories based on the principal cause of electrostriction in under electric field \cite{OrigEAP}. In one kind, electrostriction is related to the intrinsic field-induced conformational changes in the molecule, This variety is known as a \textbf{Ferroelectric Polymer Actuator}, and they have partial crystallinity and an inactive amorphous phase (e.g. Poly(vinylidene fluoride) or PDVF). In another kind, The primary cause of electrostriction is extrinsic electronic charge attraction/repulsion between two electrode layers of the actuator - This type is known as \textbf{Dielectric EAP} (Fig. \ref{fig:DEEAP}). Applied electric field creates a push/pull across those surfaces, causing the intermittent material to squish/stretch, respectively, creating a strain of 100\verb|%|. Currently, most of the studies are based on silicone elastomer based Dielectric EAPs that have been shown to actuate reliably for a large number of cycles. Self healing polymers are also considered for this variety of EAPs \cite{PolarEAP}, \cite{CurrSilic}, \cite{SelfHeal}.

               
                   


                
        \end{enumerate}

        \begin{table}[H]
        \scriptsize 
         \begin{center}
                \caption{Advantages and Disadvantages of the two types of EAPs. \cite{OrigEAP}}
        \label{tab:EAPType}   
        
       \begin{tabular}{p{0.15\linewidth}p{0.3\linewidth}p{0.45\linewidth}}
       
        \toprule
               \textbf{EAP Type}& \textbf{Advantages}&\textbf{Disadvantages}\\
             \midrule
               Ionic EAP& \begin{itemize} \item These can produce huge bending displacements  \item Require comparatively very low voltage  \item They are naturally bidirectional actuators depending on the voltage polarity \end{itemize} & \begin{itemize} \item Generally, they do not hold strain very well under DC voltage \item They have slow response rates (in the order of fraction of a second)  \item They are capable of inducing a relatively low actuation force (bending) \item In aqueous systems, ionic EAP can get electrolyzed at \verb|>1.23V| \item To operate in air requires attention to the electrolyte \item Low electromechanical coupling efficiency \end{itemize} \\
               Field-activated EAP & \begin{itemize} \item They operate stably for a long period of time without requiring much attention \item They have a rapid response (milliseconds)  \item They can hold mechanical strain under DC activation \item produces large actuation force  \end{itemize} & \begin{itemize} \item They require huge electric fields, thereby need high operating voltages. They generally require an order of 150 V/ micrometer for 10 percent strain. Recently, using composite DE is allowing for (around 20 V/ micrometer)   \item They function by electrostriction. Thereby, have monopolar actuation, independent of the voltage polarity. \end{itemize} \\
             \bottomrule

        \end{tabular} \\

        \end{center}
        \end{table}
        
         Both varieties of EAPs are compared in the table \ref{tab:EAPType}. The large actuation available at small operating voltages reflects the potential applications of Ionic EAPs in robotic prosthetics, robot-assisted surgery, etc. While Dielectric EAPs are not as much weighed for biomedical applications due to high driving voltages, they are better studied for other applications requiring large strains with a small footprint. \\

\item \textbf{Electrorheology and Magnetorheology: } 

\label{sec:ermr}
\red{Rheology is a field of physics that studies how materials flow or deform under applied forces or stresses. Rheological properties are the properties that determine the exact nature of response the material would have under such influences. These properties for Electrorheological (ER) or Magnetorheological (MR) materials are strong functions of the electric or magnetic field strength applied to them. ER and MR fluids are suspensions of micron-size particles in non-conductive carrier oil they transition from a liquid to a pseudo-solid in the presence of a magnetic field for MR fluids or an electric field for ER fluid of sufficient strength. They develop a yield stress when subjected to external stimuli \cite{RevMRER}. Some examples of these materials include polymers and gels, controllable MR and ER fluids, MR composites, MR and ER elastomers, Magnetostrictive materials and liquid crystals etc.}

\red{They have many applications such as seismic dampers, bullet proof devices, vibration damping in locomotive engine mounts, gun recoil systems, tuneable vehicle suspensions etc. \cite{RevMRER}. MR and ER actuators for robotic applications are almost entirely used in haptic systems \cite{MRFeasibility} \cite{hapticMR}. These haptic systems have been deployed, for example, in many surgical robot systems. Traditionally, robotic systems for surgery have suffered a lack of tactile feedback for the surgeon, which is crucial for precision in delicate operations. Multiple haptic master devices have been presented to solve this problem and provide haptic feedback \cite{6dofservo}. These servo-motor based actuators have compromised precision at low torques due to cogging phenomenon and brush friction. This is one of the optimal use cases for ER and MR based haptic controllers as they can provide smooth-continuous motion by changing from liquid to solid quickly. }

\red{Apart from haptics, another major use-case for ER and MR fluids is in decoupling mechanisms such as clutch. When we talk about traditional actuation mechanisms for powered robotic prosthetics, the weight of motor can be displaced using cable driven mechanisms as discussed in topic \ref{tendonmechanisms} of this section. However, the reflected actuator inertia at the linkages and at the manipulator can be huge. To soften the effect of this inertia, various methods have been proposed like Series Elastic Actuator (SEA), Variable Stiffness Actuator (VSA), Series Damping Actuator (SDA), etc. which work by decoupling the reflected actuator inertia \cite{MRFeasibility}. However, due to elastic nature of these dampers, the controllable bandwidth of the manipulator is reduced significantly. An ER or MR Fluid based clutch can be an ideal solution in this case as the torque transmitted can be controlled by the intensity of the applied field \cite{MRFeasibility}. }

\red{In general MR fluids see more application due to certain characteristic issues with ER fluids. These include narrow operating range, lower yield stress and sensitivity to impurities etc. However, lower material cost and lower density (typically 2 to 3 times, as compared to MR fluids), make them favourable for weight sensitive applications. }\\


\item \textbf{Electroactive Ceramics: } 

Electroactive Ceramics \red{in the form of piezoelectric actuators} are another atypical mode of motion generation. The word "Piezen" means to squeeze or press. The mode of actuation is again based on Electrostriction, and they have both sensor and actuator applications (Fig. \ref{fig:Piezo}). Many industrial applications are seen for piezoelectric actuators, such as Inchworm motors \cite{preskin}. Piezoelectric actuators are typically used in high precision and small displacement tasks, with limited application in the field of robotic prosthetic limbs due to their high precision and naturally microscopic range of motion. \red{One example of their application in the field of prosthetics is digital Brail tablets for the visually impaired.} \\

\textcolor{black}{Their use as a sensor is very prominent. The crystal lattice in these materials exhibits a change in shape when an external field is applied, which results in movement at the lattice level causing actuation. The same material, when subjected to force, will generate an electrical potential that is used for sensory application. These kinds of sensors are called piezoelectric sensors. Piezoelectric sensors based on electroactive ceramics are used for dynamic pressure sensing applications such as vibration and force sensing in various engineering challenges. Their application in prosthetics ranges from touch sensing to stimulating proprioception and force feedback when used in tandem with a suitable neural interface, as discussed in the control section of this paper. The general topic of Piezo-sensing and applications thereof are discussed in more detail in the E-skin column of section \ref{sec:paralleltrends} in this paper.}

One popular example of Electroactive ceramics is - PZT (Lead Zirconate Titanate), Zinc Oxide (ZnO), etc.

        \begin{figure}[H]
             \centering
             \includegraphics[width =0.6\linewidth]{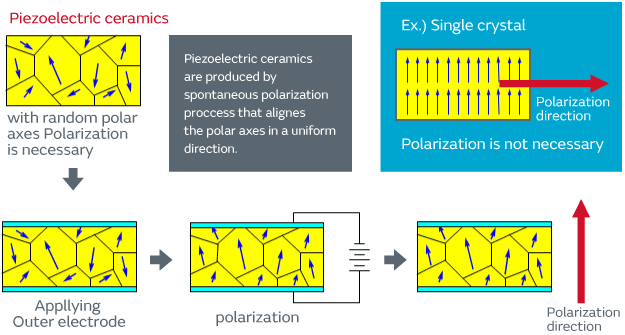}
             \abovecaptionskip=0pt
             \caption { Principle of Piezoelectric Actuators \cite{PeizoWebsite}}
             \label{fig:Piezo}
          \Description[Principle of Piezoelectric Actuators]{This figure illustrates the working principle of piezoelectric actuators where the polarization alignes the polar axes in a uniform direction.}
            \end{figure}

\item \textbf{Material Jamming: }

Material Jamming is a popular soft robotic actuator mechanism that allows for the development of novel systems with great stiffness variation and minimum volume variation. Oceanic expeditions, industrial grabbing, and usage as paws for bio-mimicking robotic mobility are just a few of the applications that have been recorded. There are primarily 3 classes of Jamming actuators: granular, layer, and fiber jamming actuators \cite{MJDemonstration}. The fundamental principle is that filler material of the required properties is enclosed within an elastomeric membrane and the introduction or elimination of fluid substrate from within due to vacuum or pumping in will respectively solidify or liquefy the whole composition. The rigidity of the system is regulated by altering the suction or by using harder filler material inside.\\


\item \textbf{Tendon Driven Mechanisms: }
\label{tendonmechanisms}
hile not technically actuators by themselves, these bio-inspired materials are widely popular in prosthetic applications. Essentially just like in an animal skeleton, the muscles transmit their force to bone through a tendon, the tendon mechanisms are used to locate the actuator away from the location of actuation. This is extremely useful in designing larger robotic systems such as a robotic arm. Motors or other actuators that will be used for force generation can be mounted away from space-constrained locations, such as fingers. Tendon mechanisms also allow us the option to distribute the forces across the actuated part, rather than concentrating them at a location – as would be the case if the actuator (say, motor) is directly attached to the end of the appendage. Force transmission through the tendon also allows some compliance to the system compared to rigid actuator attachment, as the nature of the material used for the tendon can be varied if properties such as elasticity are desired. The way tendon is routed through the machine also manipulates the actuation capabilities of the system. They may also be used for force amplification of the original actuator using a variety of methods such as pulley, hydraulic amplification, etc.

In a way, Tendon-driven systems are homologous to the traditional thread-puppet operation. A variety of complex motions are possible using tendon mechanisms, and they surely expand the scope of robotic actuators multifold. Tendon mechanisms are also considered as adapters on top of traditional hard robotic actuators such as servo motors to give them the compliance and other capabilities of soft robotic actuators. 
\end{enumerate}

A summary of the above-discussed soft robotic actuators is given in table \ref{tab:SoftRobotics}.
\scriptsize 

\begin{longtable}{|p{0.5in}|p{0.5in}|p{1.5in}|p{1.0in}|p{1.0in}|} 
\caption{Summary of types of Soft Robotic Actuators. \cite{BiomedSoftRob}}
\label{tab:SoftRobotics} \\

\hline 


Flexible Fluidic Actuators (FFAs) & - \cite{FFAHandME}, \cite{UASoftHand}, \cite{FFAElbow}, \cite{ElectricalTMM}, \cite{SelfHeal}, \cite{FluidOrigami}, \cite{SelfHealReview}, \cite{HydrogelPMM}, \cite{StrPump}, \cite{FFATyp}, \cite{NoPumpMagFFA}  & An inflatable elastomeric liner is activated by pumping liquids/gas into the atrium. The liner is encased by cleverly engineered sheathing that converts inflation of the inner membrane into an equatorial expansion with axial contraction of the overall actuator. 

Typical examples of FFAs include - McKibben Muscle, Thin McKibben Muscle, PneuNets, and PneuFlex. & Compliance and Flexibility, Variable Stiffness, High Force-to-Weight Ratio, Simple Design and Fabrication, and Active research on Self-healing and Thin McKibben Muscles are increasing the scope and practicality. & Complex behavior of fluids may require complex control systems, Speed is limited, and the Current generation of FFAs have high maintenance requirements.  \\

\hline

Shape Memory Materials & Shape Memory Alloys (SMA) \cite{SMAElbow}, \cite{SMAArm}, \cite{CoiledSMAMusc}, \cite{StressStrainSMA} & These materials have a critical transition temperature at which they are formed into a specific shape. After plastic deformation, if the material is brought back to the transition temperature, it returns to its initial geometric form. Some of these materials are also studied for their super elastic nature around the transition temperature. 
       Example: Nitinol (Ni14Ti11) & Compact and Lightweight while offering high energy density, Simple actuation Mechanism, silent operation & Limited range of motion, Slow response time, Hysteresis and Creep, Temperature sensitivity, etc. 
\\

\cline{2-5} 

 & Shape Memory Polymers (SMP) \cite{SMABook} & Shape Memory Polymers are nonmetallic analogues of SMAs. Typical compositions are Polytetrafluoroethylene (PFTE), Polylactide (PLA), ethylene-vinyl acetate (EVA), etc. & Ease of manufacturing and scaling, Customizable actuation parameters & Lower operating temperatures, lower actuation forces generated compared to alloys, limited applicability. 
\\

 \cline{2-5} 
 & Shape Memory Composites (SMC) \cite{SMAThermoStudy}, \cite{SMCAero} & Usage of mechanical engineering designs such as compliant mechanisms and composite constructions of materials with desired properties, etc. are used to complement the unilateral SMA or SMP-based actuation and achieve desired actuation. & Customizable actuation based on mechanical design. Compliant mechanisms and other mechanical engineering techniques are used to expand the scope of applicability of the Shape Memory actuators.  & All inherent limitations of SMAs and SMPs exist here also, as they are essentially the core actuators in these composites.  
\\
\hline

 Electroactive Polymer & Ionic EAP \cite{BGA25}, \cite{BGA26}, \cite{IonicEAPFab}, \cite{NASAEAP}, \cite{PolarEAP}, \cite{TempAssSACP}, \cite{IPNCPLinAct}, \cite{PolyPySPE}, \cite{DenseCNTEAP}, \cite{PdIEAP}, \cite{AgIEAP}, \cite{IonicEAPReview}, \cite{Polyaniline}, \cite{ILEAP}, \cite{ImidazoliumIL} & "Their actuation is based on diffusion of Ions. Their general structure is constructed with a film containing a suitable electrolyte, which is sandwiched by two flexible electrodes. These exhibit deformation in the form of bending which is due to accumulation of the ions closer to one electrode, causing that side to bulk up, and the other to shrink. 

Typical Examples include IPMC, BGA, NAFION-PANI Based EAP, etc."& Lower voltage operation, Easy manufacturing, and scaling, small footprint, mechanically compliant, good weight-to-force of actuation ratio. Bio-compatible varieties of these actuators are available. & Typically lower actuation forces are generated. For forces useful for prosthetic operation, complex arrays of these polymers are proposed, but that creates fabrication limitations. \\

 \cline{2-5} 
 & Field Activated EAP \cite{CurrSilic}, \cite{EAPChemRev}, \cite{OrigEAP}, \cite{WW-EAP} & These kinds of EAPs exhibit their deformation due to the principle of Electrostriction. They are also called Dielectric Transducers (DETs). The effect is independent of polarity. Their mechanism of actuation involves producing an electrical charge on the two opposing surfaces of a soft dielectric material. This creates a push/pull across those surfaces, causing the intermittent material to squish/stretch, respectively.  There are two subtypes: Dielectric EAP and Ferroelectric EAP. & High actuation force per unit weight of the actuator is the biggest merit of these actuators. The actuator design is simple, and the use of correct elastomers can result in high repeatability and long life of the actuator. they are also known to be energy efficient in their operation.  & The operation of these actuators typically requires high electrostatic potential. Such high voltages make them undesirable for use in prosthetic applications.

\\
 
 \hline

     Magneto- rheology/ Electro- rheology & - \cite{Electrorheology}, \cite{EOActuator1}, \cite{EOActuator2}, \cite{EOActuator3} & These kinds of actuators contain magnetic or electrically active particles which reorient and interact upon application of magnetic and electrical fields respectively. This principle forms the functional basis of these actuators & The ability to precisely control at small (e.g. slim haptic actuators for the Brail tablets) and large scale (e.g. dampers and clutch mechanisms) alike. Quick response with smooth and continuous motions.  & They are primarily not motion generation devices. May limit the actuator's bandwidth when used in damping applications. Temperature sensitivity, Weight, density, and cost factors might be limiting.  
\\

 \hline 
Other Actuators*: & Electroactive Ceramics  \cite{EAPLatRev}, \cite{PeizoWebsite} & "These are ceramics that show motion due to the principle of electrostriction. Their applications are usually limited to very high precision and low strain purposes due to their naturally microscopic actuation. Inchworm motors are based on these actuators and are used in powerful microscopes and other scientific devices requiring nanometer precision. 
Example: PZT (Lead Zirconate Titanate)" & As the actuation is based on the crystal lattice level movement within the material, these actuators are useful in use-cases requiring high precision and repeatability. They are also used for sensor applications, & They are extremely precise microscopic actuators with limited application in prosthetic applications. The range of motion obtained using these actuators is low.  \\

 \cline{2-5} 
 & Material Jamming \cite{MaterJamTypes}, \cite{SelStiff}, \cite{TensMatJam}, \cite{MaterialJammingPic}, \cite{JamRobots} & "The fundamental principle is that material is "jammed" into the desired orientation. One way to achieve this is by creating a pseudo fluid mass (e.g. solid granules in a pool of liquid) and removing fluidity to set in the desired shape. There are primarily 3 classes of such actuators: Granular, Layer, and Fiber Jamming Actuators"& They are simple actuators offering a high degree of compliance. Typically used for gripping applications, the gripping characteristics can be tuned based on the size, shape, and nature of the jamming material used within the actuator. & Their application is limited to a few use cases. They are not useful for generic motion generation purposes.  \\

 \cline{2-5}  
 & Tendon Driven \cite{SoRoThumb}, \cite{TendonHand19DOF}, \cite{NLTendonActuator} & "They do not qualify as “actuators” by themselves, however, they are of the paramount value for expanding the scope of the conventional actuators. They can be used to change the location of the actuator compared to the actuation site, to manipulate the magnitude, direction, and nature of force applied by the actuator, to add compliance to the system, etc." & Tendon mechanisms expand the applicability of conventional actuators by providing compliance to the actuator and the ability to mount the heavy core actuator away from the site of application.   & All limitations of the core actuator are reflected here also. Increased backlash and reduced repeatability might result based on the material used for the tendon. 
\\
 
\hline 

\hline
    
\end{longtable}

*Other Actuators: These are not soft robotic actuators by definition, but are being included in the discussion for their interesting properties and use cases.
\normalsize


\section{Control systems for Robotic Prosthetic Limbs}
\label{sec:control}
A robotic prosthetic arm or a leg has to be an extension of one’s human stature. The hardware has to function on its human-ware’s command in an organic way. While studies into developing a viable mechanical design and motor system for robotic prosthetic limbs are important, the limb itself is useless unless there is a sufficiently streamlined way to control the limb. An ideal control system should be able to en route sensory and motor signals bidirectionally. It should be able to convey the state of surroundings from the prosthetic limb to one’s human consciousness, and be able to convey one’s intentions to the prosthetic limb. This section will cover some of the research areas currently active for the development of techniques that can enable the control of robotic prosthetics. A few of these options have been well explored and documented in the literature, while some options remain under active research for now. Beyond the purview of sensing the user’s intentions, Prosthetic control can be of two types: Proportional and Pattern-based. Proportional control gives the user control over position, velocity, and force of actuation over the chosen DoF of the prosthesis. However, constant feedback is required for proportional control of a prosthetic limb (Closed-loop control). Pattern-based control is where the motion of the prosthetic limb is predefined and is executed based on the classification of user intention using the sensor data. The optimal control strategy in most scenarios is a tailored combination of proportional and pattern-based control. \cite{MyoelecControl} presents the designing of control strategies for active prosthesis control.

 \begin{enumerate}

   \item \textbf{Residual Muscle Contractions: } It is noted that when it comes to controlling an externally powered prosthetic limb, technical failure can be detrimental to the user and people around them. So control strategies for such prostheses are based on proven and preferably simpler technologies that will be reliable at their task. This is the reason that most of the research in control strategies for robotic prosthetic limbs are concentrated around trans-radial and trans-tibial applications by measuring activity in muscle remnant after amputation. \cite{ReviewControl} There are numerous techniques covered in the literature that elicit a variety of principles for measuring muscle activity, the table \ref{ResidualMuscle} covers some of the research.
 \begin{itemize}
     \item \textit{Electromyography (EMG): } EMG is the most prominent control strategy for robotic prosthetics based on measuring muscular contraction. Muscle acts as an amplifier for the motor neural signals coming in from the brain. These signals can be read through EMG. Such readings can be taken either from the residual part of the amputated limb, or from the base of the limb. While the latter is more complex to interpret into usable signals, there have been studies to make meaningful interpretations of these readings so that they can be used to control a robotic prosthetic. Based on the type of collection technique, EMG can be divided into two types: Surface EMG (sEMG) and Intramuscular EMG (iEMG). In sEMG, the electrodes are placed on the surface of skin \cite{sEMGluca} \cite{SurfaceEMG} - This limits access to the superficial muscles only. The data obtained from sEMG is also inferior to iEMG due to the distance between actual muscle actuation and skin surface. Therefore, usually, multiple electrodes are used and the clarity of signal will depend on the number of electrodes employed. iEMG on the other hand required implantation of electrodes either under the skin \cite{iEMG} or inside the muscle (needle electrodes \cite{iEMG1}). This gives better data for controlling, however requirement of invasive implantation is a major demerit.
     \item  Myoelectric (controlled by electromyography, EMG) prostheses are currently available for trans-radial amputees (who make up the majority of arm amputees). Myoelectric prostheses have the potential to improve the quality of life for arm amputees. Impending progress in the control systems is one of the major reasons for their low adoption. Work published in \cite{OnlineEMG} proposes a protocol for developing and testing algorithms for force control and movement recognition on a benchmark database. This paper along with the majority of others in this field is dependent on the availability of prepossessed data. Online algorithms have been proposed for these control strategies since as early as 2008 \cite{EMGHand}. However, as the degree of freedom increases, the efficiency of these algorithms declines substantially. So far, up to 4 degrees of freedom have been successfully demonstrated through online algorithms. Reference \cite{EMGHand} reviews the recent trends in the control strategies of robotic prosthetics using EMG controls. Implanted Myoelectric Sensor Systems also exist that provide better accuracy \cite{IMES}. 
 \end{itemize}

       \scriptsize 
            
       \begin{longtable}{|p{1.0in}|p{2.25in}|p{1.25in}|p{0.5in}|}

        \caption{Residual muscle-based control strategies. \cite{ReviewControl}}
       \label{ResidualMuscle} \\
       
        \hline
               \textbf{Technique}& \textbf{Description}& \textbf{Remarks}& \textbf{Reference}\\
             \hline
              Electromyography (EMG) & Measuring electrical potentials generated during muscle contraction. The frequency range is up to 500 Hz, with a peak amplitude of 1 to 10mV in the 50 to 150 Hz range. EMG can be of two types, Surface EMG (sEMG), Intramuscular EMG (iEMG). & Most popular technique for prosthetic limb control.
 & \cite{Ossur}
 , \cite{ReviewControl}
 \\

      \hline
              
               Magnetomyography (MMG) & MMG are conceptually similar to EMG, with the difference being that here the subject is the low amplitude magnetic fields produced by the electric current propagating through muscle during contractions. & The fields observed have their amplitudes in the range of pico-tesla and femto-tesla ranges and are prone to pollution from environmental magnetic noise. & \cite{ReviewControl}, \cite{MMG1} 

            \\
             
      \hline
             Optical myography (OMG) & This technique is based on a camera looking at the surface of the skin to track minute deformations caused by muscle movements. Optical markers are often placed on the surface of the skin to enhance contrast, which aids in better tracking by the camera. & Deep muscle movements are not trackable by this technique, and it is extremely difficult to implement if there is limb movement. & \cite{OpticalMG}
 \\
             
      \hline
             
           Sonomyography (SMG) & Morphological changes in muscles are recorded using ultrasound waves generated by piezoelectric transducers that are reflected from tissue. The acoustic impedance of the tissue will affect the reflected signal properties. & Typically, requires bulky equipment. Readings are very sensitive to the correct positioning of the detector. &  \cite{SMG}, \cite{ReviewControl}
 \\
             
      \hline
             Force Myography (FMG) & Force sensors around limbs for detecting dimensional changes in the superficial muscle groups.  & Most prominent control strategy for robotic prosthetic limbs. & \cite{WristFlex}
 \\
             
      \hline
              Phonomyography (PMG) & PMG are based on the low-frequency oscillations exhibited by muscle fibers during contractions. The vibrations usually range between 5 and 100 Hz and are detected using microphones or low-mass accelerometers placed over the skin on top of the muscle.  & No Interference, Usually used along with other techniques. &  \cite {HybridPMG}
 \\
             
      \hline
             Electrical Impedance Tomography (EIT) & Electrical impedance of the body part are measured using an array of surface electrodes. Signals are in the 10 KHz to 1MHz range. Common types: 2 Terminal, 4 Terminal  & Popular uses include Blood flow study, Brain activity and  Lung ventilation monitoring, Hand gesture recognition, etc.  & \cite{ReviewControl}
 \\
             
      \hline
             Capacitance Sensing & This technique is based on the capacitance between the sensor and the object and, similar to EIT, it also uses transmitter-receiver pair electrodes. & More sensitive to subtle movements in the muscle and surrounding tissue to appreciable depths under the electrode array. Prone to parasitic capacitance and steep sensitivity to a person's physiology. & \cite{CapPiezoTandem}
  \\
             
      \hline
             Near IR Spectroscopy (NIRS) & Tissues in the human body are quite transparent to the near IR range of the electromagnetic spectrum, i.e., 700 to 900 nm. IR light emitted into muscle tissue is scattered around and detected by a detector. The depth of NIRS signal can be changed by changing the distance between the detector and the emitter. & Prolonged usage may heat up the tissue. & \cite{NIRS}
\\
             
      \hline
             Optoeelectronic Myography & Functionally similar to NIRS with one difference being the usage of red light instead of IR, preventing heating issues associated with IR.  & Shortfall is that red light has lower penetration compared to IR. & \cite{OptoelecMG} 
 \\
             
      \hline
             Cineplasty & A surgical procedure where a mechanical link is created between residual muscle and the prosthetic limb. The original concepts date back to the 19th century by Vanghetti. Fixtures such as Bowden cable were used for the process. The procedure is no longer popular due to the detrimental anatomical effects of the procedure and infection. The force generated through such a linkage was also small. Recently, experimental approaches involving electromechanical amplification in cineplasty are being considered (Extended Physiological Proprioception (EPP)) & One commendable achievement of such a rudimentary control strategy that most others in today’s area fail to achieve is proprioceptive feedback. This is due to the retention of the natural agonist-antagonist muscle connections. & \cite{Cineplasty}, \cite{ReviewControl}
 \\
             
      \hline
             Myokinetic Control Interface (MYKI) & Permanent magnet markers are implanted into remnant muscles. The fundamental concept is that localizing the markers is equivalent to measuring the movement in muscles in which it is implanted since the magnet moves with muscle. & Both Isometric and Isotonic contractions of muscle can be measured. It is another control strategy with the potential to restore proprioception. & \cite{MYKI1}, \cite{MYKI}
 \\
 
             \hline

        \end{longtable} 
         
      \normalsize

    \item \textbf{Targeted Muscle Reinnervation: } Targeted Muscle Reinnervation is a surgical technique where the nerve endings of the residual limb are relocated to a site different from the area of amputation to reinnervate muscles at the target muscle site, where these damaged nerve endings regenerate and attach to their nearby tissue. If the nerve endings are not attached to tissue, they may grow into a bundle and develop neuroma. In case of amputation, TMR prevents neuroma and phantom limb pain while also providing an anatomical control site for a prosthetic limb. The re-innervated muscles amplify the motor neural signals, which can then be captured through EMG and used for control of the robotic prosthetic limb. Figures \ref{fig:TMRPoster}, \ref{fig:TMR1} depict the TMR - EMG-based control strategy for a robotic prosthetic arm. 
Some of the initial literature about TMR was published by Kuiken et al. \cite{TMRImage2} \cite{TMRImage1} where they cover the surgical procedure as well as sensory mapping of the nerves. The paper \cite{TMREMG} goes into detail about the signal processing for such a prosthetic arm. TMR provides a way to facilitate near-natural control of full-length arm and leg amputations. The paper \cite{TMR1} covers a study where a 41-year-old trans-humeral and shoulder disarticulation who had undergone TMR during amputation was able to control robotic prostheses with the technique. \cite{TMR} covers the design of an EMG-based optimal control algorithm for TMR amputees. TMR is a very feasible mode of control for a robotic prosthetic. TMR can be performed during or after amputation in a delayed setting.  

           

 \begin{figure}
     \centering
     \begin{subfigure}[b]{0.45\linewidth}
             \lineskip=-\fboxrule
             \centering
             \includegraphics[width =\linewidth]{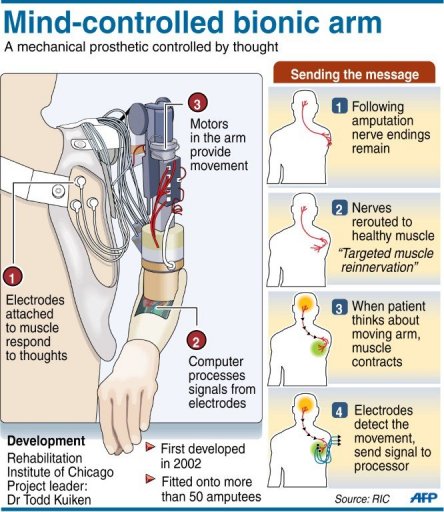}
             \abovecaptionskip=0pt
             \caption { Bionic arm control illustration using TMR and EMG. \cite{TMRPoster}}
             \label{fig:TMRPoster}
          
            \end{subfigure}
            \hfill
    \begin{subfigure}[b]{0.4\linewidth}
        \centering
        \includegraphics[width=\linewidth]{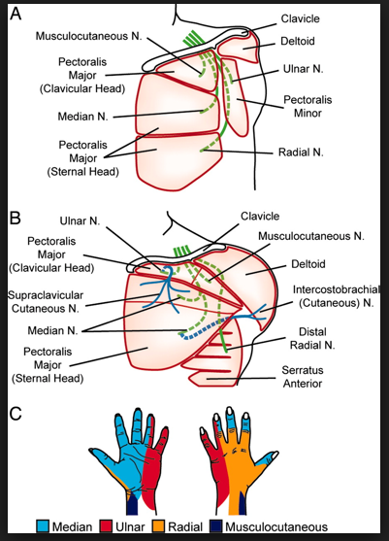}
        \caption{ Concept illustration for TMR for arm amputation\cite{TMRImage1}}
        \label{fig: TMRConcept}
     \end{subfigure}
     \hfill
    \hfill
        \caption{ Targeted Muscle Reinnervation}
        \label{fig:TMR1}
        \Description[Targeted Muscle Reinnervation]{This figure illustrates the working principle of Targetted Muscle Reinnervation through a popular poster. The subfigure (b) indicates an example of TMR application discussing various nerves responsible for controlling a hand and their potential reinnervation sites.}
\end{figure} 

    \item \textbf{Neural Implants and BCI: } As discussed in Section II: Neuroprostheses, neural signals can be artificially fetched (Fig. \ref{fig:NeurralImplants}). The collected signals can be processed to control an external prosthetic limb. Or these signals can be used electrically to stimulate paralyzed body parts for restoring function. This technique is called Functional Electrical Stimulation (FES).  There is a lot of potential in combining BCI to FES in people with partial paralysis \cite{BCIApp}. Pfurtscheller et al. \cite{FESArm} created a program that allowed a patient with a spinal cord injury to use his paralyzed hands to grab a cylinder. The patient imagined moving their feet to create beta oscillations, which were then recorded with EEG, categorized by the BCI, and utilized to operate the FES device, which caused the hand's muscles to contract. One of the recent innovations in this field is small injectable electronic modules – called BIONs (BIOnic Neurons) - which get implanted in motor neurons at several sites in the body, which read the neural signals and execute FES to reanimate paralyzed body parts given by Tan and Loeb in 2007 \cite{BION} \cite{BION1}. An external radio frequency coil supplies power and digital command data, and they use stimulating current pulses to attract motor neurons and trigger corresponding muscles. Neural implants have been studied to be promising solutions for control and sensory interface between the human body and prosthetic limb, However, the invasive nature of such an interface and the associated risks are the major problems. Recent work published by Neuralink \cite{Neuralink}  has been motivating towards an implanted neuroprosthesis-powered Brain-Computer Interface capable of controlling robotic prosthetics. The package of over 3000 electrodes in a package the size of a penny is commendable. The device will allow minimally invasive access to numerous individual neurons or neural clusters, which is very promising. The fact that there are private and public entities working towards standardization of equipment and procedures towards BCI technology development is a commendable leap for humanity and will help us in the future to counteract all sorts of impairments and ailments with technology. 
    
    \begin{figure}[h!]
    \lineskip=-\fboxrule
     
    \begin{subfigure}[b]{0.52\linewidth}
        \centering
        \includegraphics[width=\linewidth]{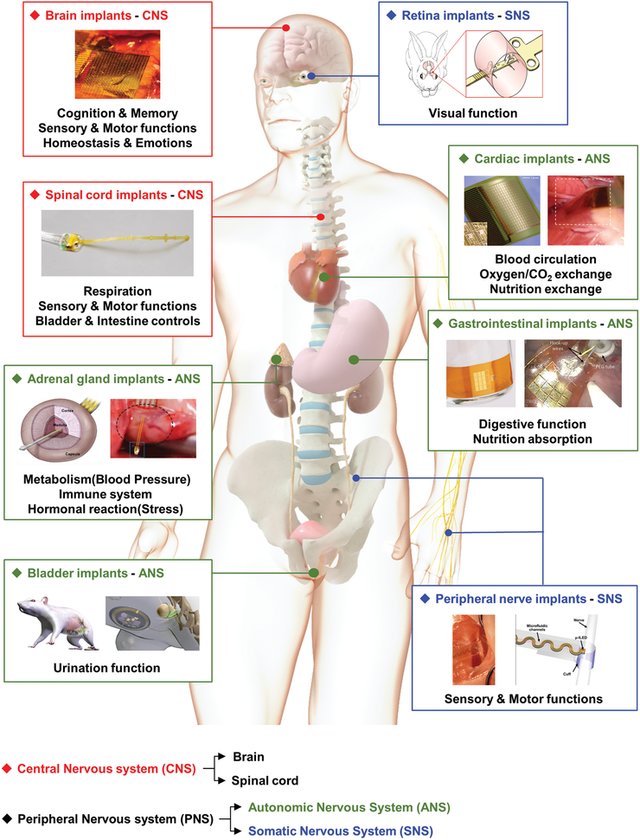}
        \caption{Current overview of neural implants applications\cite{TMRImage1}}
        \label{fig: NeuralImplantsTypes}
     \end{subfigure}
     \hfill
     \begin{subfigure}[b]{0.38\linewidth}
        \centering
        \includegraphics[width=\linewidth]{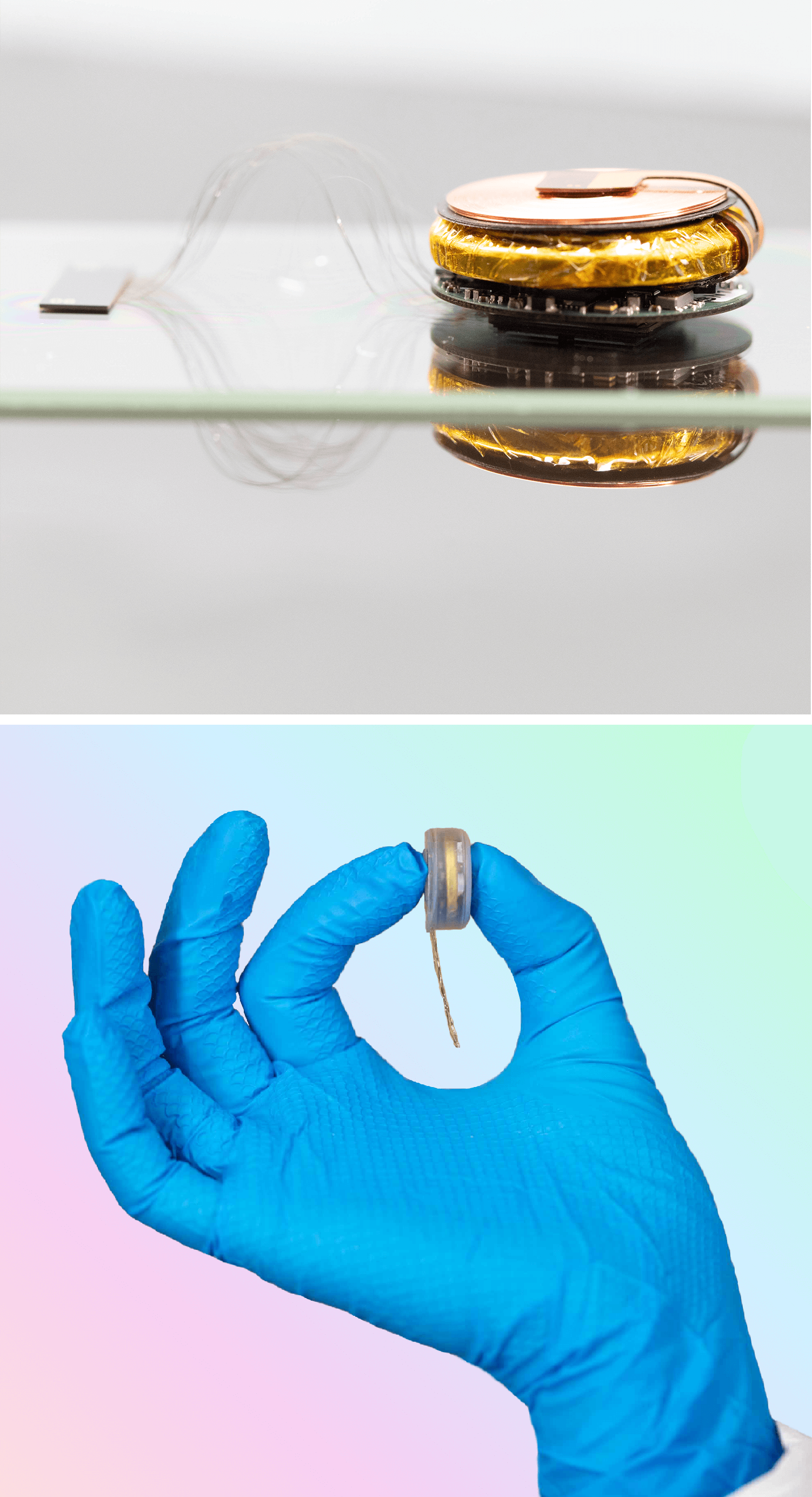}
        \caption{Microrobot - BCI Chip \cite{Neuralink} \cite{NeuralinkImg}}
        \label{fig:neuralink}
     \end{subfigure}
    \hfill
        \caption{\textbf{Neural Implants } }
        \label{fig:NeurralImplants}
        \Description[Neural Implants]{This figure shows a broad overview of the applications of neural implants. The subfigure (b) shows a BCI neural implant held in a person’s hand for scale.}
    \end{figure} 
      
    \item \textbf{Electroencephalography (EEG) and BCI: } EEG is a non-invasive way to detect the brain's neural activity. An EEG signal is fetched from outside the skull as compared to signals fetched through Brain Implants. Mostly, the signals received from EEG have low amplitude and low resolution. Fetching signals from individual neurons with EEG is not possible with the current level of technology. Electroencephalography data obtained is a superimposition of thousands of brain waves representing brain activity, and active research is underway to transform the superimposed waves to their individual components \cite{Neurosky}. Being able to separate and interpret these individual neural signals means being able to directly read one's intentions from the brain itself.  The application of Electroencephalography for control of a robotic prosthetic is a novel idea. This can be the ideal solution once we're able to understand and interpret the minutes of EEG signals, which is a complex task and a far stretch right now. As of now, low-resolution interpretation of these signals (brain waves) has been studied and researchers have been able to characterize these signals into a few types ($\delta$ , $\theta$ , $\alpha$ , $\beta$ , $\gamma$). This interpretation has been demonstrated to be useful for low DoF control strategies for robots. A famous demonstration of this concept is thought-controlled toys.\cite{ThoughtToy} 

   







    Active research is underway in this field and future prospect of its application are enticing. The process of BCI control and its sections are explained in detail in the paper: \cite{BCIRobpro}. Recent work demonstrates the use of BCI for controlling a Robotic Prosthetic Arm \cite{BCIApplication}. Inherently, such a task requires high-dimensional and real-time control from the EEG Signals. Consequently, it has proven challenging to fully manipulate robotic prosthetic limbs with dexterity using a BCI system. But such functionality is necessary if BCI-controlled robotic or prosthetic limbs are to be used outside of research laboratories. \cite{BCIApplication} reviews the current trends in addressing these problems. With the introduction of affordable and easy-to-operate Electroencephalography sensor kits, research in this area has accelerated and more and more studies are being brought into the mainstream. a 2014 publication by Katona et. al. Traditionally EEG research was limited to big neuroscience labs only, however recently, a few affordable and simpler EEG kits have been introduced into the market which is surely helping research into this field. \cite{Neurosky} discusses the application of one such commercially available and affordable electroencephalography kit called Neurosky MindFlex\textsuperscript{\textregistered}
    
    \item \textbf{Other Human Machine Interfaces (HMI): } In many cases of motor disability, Other control strategies are not viable due to a multitude of reasons. In the current times when BCI technology is still in its primitive stage, for patients with complete motor disability, the aim for assistive technologies is usually limited to simpler tasks such as facilitating personal movement from one place to another or enabling communication, etc. Some of the concepts used for control of such assistive machines are tongue movement \cite{TongueControl} (Fig. \ref{fig:TDS}),  controlled air movement through the mouth - a technique called Sip and Puff \cite{SipPuff}, voice command, face tracking \cite{HMIReview} and eye tracking \cite{ETIntAI} etc. Applicability of such control strategies in robotic prosthetic limbs with multiple DoF (degrees of freedom) is limited, but these techniques are very helpful for other assistive devices for patients with complete motor disabilities, such as in patients with Cerebral Palsy. \cite{SipPuffJared} covers an article where a Sip and Puff-based computer control system (functional as a communication prosthesis, as discussed in section 1). The system has enabled a C.P. patient named Jared to be empowered and gain some independence with technology. \cite{SnPAIWheelC} illustrates an AI-powered wheelchair control system designed for patients with complete motor disabilities. The user's intentions are collected via the Sip and Puff system for autonomous control of the wheelchair. Georgia Tech has demonstrated a tongue control system for driving control of a wheelchair \cite{TongueControlWC}. Tongue Driven Systems (TDS) are emerging into the mainstream for mobility assistance of highly disabled personnel. \cite{Tongue1} \cite{Tongue2} Typically, a magnetic marker is implanted into the tip of the tongue. The location of this magnet is tracked either using an External TDS (eTDS) headset (based on hall sensor) or an Intraoral TDS (iTDS) (based on 3-axis magnetometer sensor) \cite{TypesTDS} \cite{iTongue}. In some of the recent literature, it has been revealed that the sensory capabilities of the nerve-dense tongue tissue have the potential to provide grounds for more complex prostheses. J. Williams et al. from the University of Wisconsin \cite{TongueMonitor} have successfully demonstrated a tongue sensitivity-based 12 x 12-pixel visualization prosthesis for completely blind candidates. The device has an array of 144 electrodes which are connected to a computer through a controller. The system was able to simulate the computer display signal in low resolution (12p) to the user through electrical stimulation on the tongue surface. 
    
              \begin{figure}[H]
     \
     
    \begin{subfigure}[b]{0.3\linewidth}
        \centering
        \includegraphics[width=\linewidth]{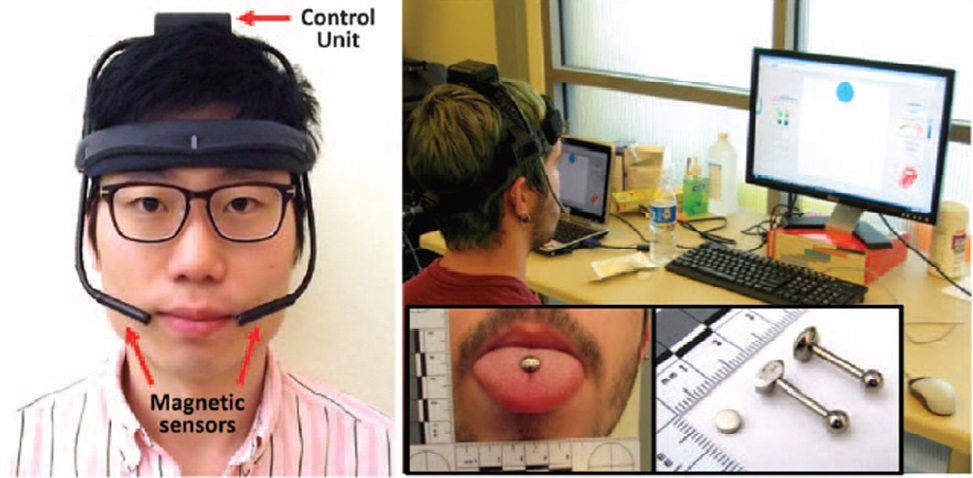}
        \caption{eTDS Demonstration \cite{eTDS}}
        \label{fig: eTDS}
     \end{subfigure}
     \hfill
     \begin{subfigure}[b]{0.3\linewidth}
        \centering
        \includegraphics[width=\linewidth]{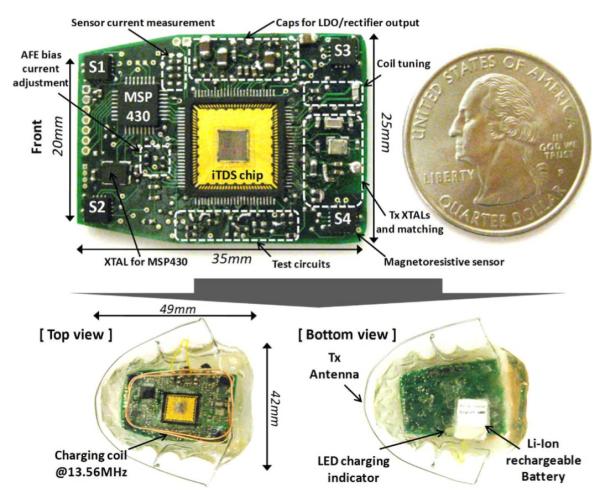}
        \caption{iTDS dental retainer \cite{TypesTDS}}
        \label{fig:iTDS}
     \end{subfigure}
     \hfill
     \begin{subfigure}[b]{0.3\linewidth}
        \centering
        \includegraphics[width=\linewidth]{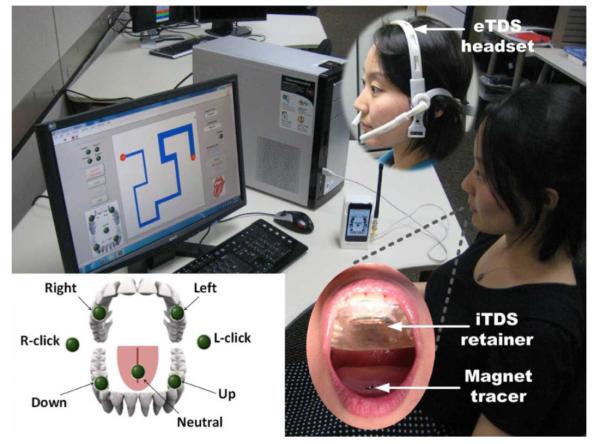}
        \caption{iTDS Demonstration \cite{TypesTDS}}
        \label{fig:iTDS1}
     \end{subfigure}
     \hfill
    
        \caption{\textbf{TDS } Tongue Driven System - HMI for prosthesis control}
        \label{fig:TDS}
        \Description[Tongue Driven Systems]{This figure shows various form-factors of tongue driven systems for prosothetic control.}
        \end{figure}
      
    \item \textbf{Integrated AI - Combining distributed sensory bank with machine learning for controlling prosthetic limbs: } Intelligent systems that collect data from a variety of sensors and use AI such as Reinforcement Learning for deciding control strategy for robotic prosthetics are under active research. This technique can be better described through an example: \\

    \textit{Consider a full-length arm amputee, wearing a robotic prosthetic limb. The intentions of this user can be interpreted for valid decision-making on the machine's end. Suppose the person is thirsty, a multitude of sensory data can be used to deduce this - for example, a basic EEG can suggest discomfort due to thirst, and dehydration can be interpreted through sensory data associated with, say blood flow or skin moisture level etc. and can be reinforced by other sensors (Say the person's body temp is slightly above normal while the motion sensors on other limbs suggest that this rise in temp. cannot be associated with a recent physical excursion): Conclusion of this analysis can be that the person is thirsty and may have intentions to drink water. Other sensors, such as cameras, etc. can be used to monitor and analyze the ambiance. If there is, say a water bottle around. The machine can associate the above analysis with decision-making as - "grab the bottle of water". The decision can be confirmed by the user through some suitable response before execution of pattern-based control of the robotic prosthetic arm for execution of the user's intentions.} 
    
    While the aforementioned is a far-fetched expectation from such a system, things like reading data from the other limbs for an arm/leg motion associated with walking/running can be comprehended by the machine more feasibly and the machine can then relay motor signals to the robotic limb to complement normal walking/running.\par 
    These systems may represent control systems of future robotic prosthetic limbs. A variety of sensors may be used to know the whereabouts of the user, these may include:
    \begin{itemize}
        \item Accelerometer-gyroscope sensors on the other limbs to understand the user's posture and intention
        \item Sensors for monitoring vitals of the user such as the amount of hydration in blood, blood glucose level, skin hydration level, etc.
        \item EEG for a broad categorization of mental state
        \item Sensors such as cameras and mics, for acquisition and interpretation of data from ambiance.
     \end{itemize}
\end{enumerate}

It is evident that the innovations in Neuroprosthesis, and Artificial Intelligence of hardware as well as software nature are generating viable options for connecting man and machine in a better way. The above discussion on control strategies for robotic prosthetics is summarized in Table \ref{tab:controlstrategies} of this paper. \\


\scriptsize
\begin{longtable}{|p{0.5in}|p{0.5in}|p{1.5in}|p{1in}|p{1in}|} 

\caption{Summary of various control strategies for robotic prosthetic control.}
\label{tab:controlstrategies} \\

\hline 
\textbf{Technology} & \textbf{Important Subtypes}  & \textbf{Description} & \textbf{Pros} & \textbf{Cons}\\

\hline
             
Residual Muscle Contraction and Targeted Muscle Reinnervation (TMR) & \cite{ReviewControl}, \cite{sEMGluca}, \cite{SurfaceEMG}, \cite{iEMG}, \cite{OnlineEMG}, \cite{EMGHand}, \cite{IMES}, \cite{TMRImage2}, \cite{TMRImage1}, \cite{TMREMG}, \cite{TMR1}, \cite{TMR} &  Control signals for the prosthetic are taken from the residual muscles of the limb. The technology used for obtaining the control signals may be various and are summarized in the table \ref{ResidualMuscle} of this paper. TMR is an extension of this technology, wherein the nerves responsible for controlling the limb getting amputated are re-routed to less active muscles of the body. This provides grounds to control a prosthetic from these re-innervated muscles. & Relatively very straightforward and is generally reliable. Realistic control of individual fingers is possible. Greater adaptability due to organic user interface. Simulated proprioception and kinesthetics are also added benefits. & Higher levels of dexterity require heavy computation and currently require per-user tuning and training of control models. Applicability is lesser in full limb amputations.  \\ 
\hline
Neural Implants and BCI & \cite{BCIApp}, \cite{FESArm}, \cite{BION}, \cite{BION1}, \cite{Neuralink} & Implants are used to electrically interface with nerves for either reading signals from or writing signals to these nerves. Thus, allowing for sensory-motor extension of the body and facilitating functional prosthetics that can feel. The technique is also helpful in certain cases of remobilization of paralyzed limbs also. & The ability to interface directly with the nerve bundle means that prosthetic control as well as sensory feedback can be facilitated in the most realistic way. & This is an invasive technique. Surgical implantation and the repercussions are undesirable to many, thus, affecting the adaptability of this technique. \\
\hline
Electro- encephalography and BCI & \cite{Neurosky}, \cite{ThoughtToy}, \cite{BCIRobpro},  \cite{BCIApplication}, \cite{BCIApplication}, \cite{Neurosky} & Electromagnetic waves generated due to brain activity are captured, and the patterns are analyzed using complex computational algorithms. Certain signals are generated by specific brain activity, and the control strategy is based on being able to decode signals specific to the limb motion. & The non-invasive nature of this technique is desirable. The strides in machine learning and artificial intelligence are greatly benefiting the applicability of EEG-based BCI for prosthetic control. & Typically, only control signals are obtained, and sensory feedback is not possible. The technology is growing, and considerable development is required to obtain practical applicability.   \\
\hline
Other HMIs & \cite{TongueControl}, \cite{SipPuff}, \cite{HMIReview}, \cite{ETIntAI}, \cite{SipPuffJared}, \cite{SnPAIWheelC}, \cite{TongueControlWC}, \cite{Tongue1}, \cite{Tongue2}, \cite{TypesTDS}, \cite{iTongue}, \cite{TongueMonitor} & These are typically lower tech methods which can safely and reliably provide some basic motor controls to majorly disabled people.The majority of the time, these control techniques are applied to powered wheelchairs and other mobility assistance equipment as well as communication prostheses.  & The basic nature of the technology allows for reliable performance, which is crucial since the goal is to provide some level of independence and control over basic activities of everyday life to severely disabled people. & The dimensionality of control is limited, and the technique is not applicable for the manipulation of complex prosthetic devices such as limbs. \\
\hline
Integrated AIs & - & This is an advanced feat of technology that is based on the collaboration between a human and a computer for decision-making and subsequently 
for prosthetic control. & The complex nature of this technology provides the potential for highly dexterous control of prosthetics. While the technology is still in its budding stage, It is the potential future of augmented reality for a lot of use cases. & The technology is in its elementary form and warrants more control with the computer. Several ethical and moral challenges have to be overcome simultaneously to the development of the technology itself.  \\
\hline

\end{longtable}

\normalsize



\section{Parallel Trends:}
\label{sec:paralleltrends}

The field of prosthetics is vast and there are many interesting topics that are beyond the general tone of this paper but are still worth adding to the discussion. The direction of these topics that will be covered are complexities of limb amputation and biomimicry attempts to restore the complex sensation of touch and proprioception.

\subsection{Proprioception}
Proprioception, also known as Kinaesthesia is our body’s ability to sense and be aware of the position and movement of various body parts. \cite{Proprioception} Proprioceptors are sensory receptors in our body that are present in all of our muscles, tendons, and joints. The brain processes the messages sent by these receptors in tandem with visual, vestibular, and other neural signals to perceive the state of your body parts – their location, position, force, torque on joints, and other characteristics of the motion. One is able to walk without consciously thinking about the placement of the next foot, or to touch one’s elbow or nose with eyes closed.  Lack of proprioception leads to things like falling while walking across uneven terrain, not understanding the own strength for example while handling objects with hands, uncoordinated movement or balance issues, etc. 
Despite all the advancements in human-mechatronic interactions in modern robotic prosthetics, Technologies attempting to assist in proprioception with prostheses are limited. As mentioned in the residual limb-based control strategies discussed in section \ref{sec:control}, Cineplasty \cite{ Cineplasty} and Myokinetic Control Interface (MYKI) \cite{MYKI}  are the two methods majorly studied for enabling kinaesthesia in trans-radial prosthetic limbs. Apart from these, AMI (Agonist-Antagonist Myo-neural Interface) \cite{AMI} is another method recently published for enabling proprioception in the transtibial prosthesis. All of these techniques attempt to re-establish proprioception in the amputation of limb extremes. Active research is underway in the area.

\subsection{eSkin}
\label{sec:eskin}
Moravec’s paradox states that what is easy for a human is complicated for a computer/robot, and usually what is difficult for a human is easy for a computer/robot. What it means is reasoning, contrary to traditional assumptions, needs a small amount of computation, however, perception skills and sensorimotor tasks require big computational resources. This paradox is very evident when we try to create artificial skin. Human perception of touch originates from a huge number of receptors (mechanical, thermal, pain) non-uniformly distributed in our skin. The sheer number of such sensor data required for efficient somatosensation in an artificial skin reveals this complex problem in its full glory. \\

In an animal body, skin and the related sensory inputs are crucial feeds for us to safely manipulate and explore the physical properties \cite{phypropeskinzx} of objects around us and for the perception of self-awareness \cite{saeskin}. In lieu of these crucial inputs, we become clumsy and unreliable in interacting with our surroundings. While artificial touch sensors of different kinds (resistive \cite{reseskin}, capacitive\cite{capeskin}, piezo-resistive \cite{preskin}, optical \cite{opteskin}, piezoelectric \cite{peeskin}, acoustic \cite{acousticeskin} etc.) in various configurations\cite{confeskin} are available, the problems in creating electronic skin are fourfold; The sensors and the embedding platform has to be compliant and soft and be capable to sense the touch in an analog manner\cite{softeskin}, A large number of such sensors are required, the large amount of tactile data generated by these sensors must be reliably transmitted, and this huge amount of data needs to be processed for sensory utilization by the robot.  \\

Researchers are now orienting their studies towards finding neural-like hardware for E skin. The large amount of data must be processed in the way these receptors work in our PNS (Peripheral Nervous System). The large number of data that needs to be transmitted approaches practical limits of communication bandwidth possible, therefore to facilitate efficient data handling, distributed low-power electronic hardware is required for computing. This system would be analogous to how PNS compliments our CNS (Central Nervous System). \cite{electronicskin} reviews the various approaches toward the development of eSkin.

\textcolor{black}{Active robotic prosthetics can already mimic many of the mechanical properties of biological hands and legs, but adding skin-like sensory capabilities could improve their acceptance and utility among amputees \cite{ProstheticSkin1}, \cite{prostheticSkin3}. The benefits of incorporating tactile feedback into prosthetic limbs are multifold, It helps in Promoting a sense of ownership \cite{ProstheticSkin}, Alleviating Phantom limb pain \cite{ProstheticSkin4}, Allowing for more natural grip control and facile operation \cite{ProstheticSkin2}, \cite{ProstheticSkin5} and proprioception \cite{ProstheticSkin6}.}

\red{The sensory receptors in human skin can be classified into seven general types \cite{ProstheticSkin}: receptors for pain, cold, warm, and four mechanoreceptors that measure innocuous mechanical stimulus. These receptors encode and convey information as action potentials (i.e.. the time between voltage spikes) and based on these action potentials, the four mechanoreceptors are studied as Slow adapting receptors (SA-I and SA-II) and Fast adapting receptors(FA-I and FA-II). Some details about these mechanoreceptors are given in the table \ref{tab:mechrecep}.}
\begin{table}[]
    \centering

\scriptsize

\begin{tabular}{|p{0.1\linewidth}|p{0.2\linewidth}|p{0.2\linewidth}|p{0.1\linewidth}|p{0.3\linewidth}|} 

\hline
\textbf{Type} & \textbf{General Features} & \textbf{Receptor} & \textbf{Class} & \textbf{Specific features} \\

\hline

Slow  Adapting Receptors  & They produce sustained signals in response to sustained stimulus. They have a slow adaption rate. & Ruffini Organs & SA-I & Located on the surface of the skin.   Provide high-resolution information about the texture and shape of the object  \\

\cline{3-5}

  &   & Merkel Disks & SA-II  & Located deeper within the skin, responsible for measuring skin stretch and thereby proprioception.\cite{ProstheticSkin} \\

  \hline
Fast Adapting Receptors / Rapid Adapting Receptors & They respond to dynamic forces and vibrations. They have rapid adaption rates and are insensitive to static forces & Meissnerr's Corpuscle & FA-I /RA-I & located closer to the skin. Lower frequency operation (5-50 Hz). Typically associated with object manipulation   and texture determination \\

\cline{3-5}
 &   & Pacinian Corpuscle  & FA-II / RA-II & They measure high-frequency vibrations (up to 400Hz) over large areas and are important for slip detection and texture detection. They are present in low density.\\
 
 \hline
\end{tabular}
\caption{Summary of different types of mechanoreceptors in human skin.}
\label{tab:mechrecep}
\end{table}

\red{The artificial tactile sensor initiatives are generally based on touch/pressure detection, where sensors for other stimuli such as temperature, strain, and humidity may be integrated. Artificial tactile feedback systems have been experimented with in prosthetic limbs as early as 1974 \cite{ReviewFlexTactHIS2}.}
\red{Conventional tactile sensors are insufficient to replicate the perspective features of human skin with flexibility, stretchability, high sensitivity, high spatial resolution, and rapid response time. \cite{ReviewFlexTactHIS}. There are primarily three types of developments in the scope of artificial skin that the research is primarily focused on: \cite{ReviewFlexTactHIS}}
\begin{enumerate}
\color{black}
    \item Multisensory systems for simultaneously detecting strain, temperature, humidity, pressure, etc.
    \item Self-powered sensors or flexible energy storage devices to realize self-sustainable tactile-sensing systems.
    \item Wireless communication modules to enable signal processing and data transmission in real-time. 
\end{enumerate}
\red{The most prominent area of current research along with making artificial skin is in the transduction mechanisms of tactile sensors. There are four categories for the development of tactile sensors in addition to some other niche sensors.}
\begin{enumerate}
\color{black}
    \item Resistive tactile sensors comprise active material sandwiched between two opposing electrodes or stacked on a pair of in-plane electrodes. The external pressure results in a change in the electrical resistance of the sensor. These sensors generally feature. These sensors feature high sensitivity, simple device structure, and facile fabrication processes at the expense of higher operating power consumption as a major drawback. Examples of resistive tactile sensors for use in various dimension applications are \cite{ReviewFlexTactHIS}: Metal Nanoparticles 
    (0D), Metal Nanowires 
    and Carbon Nano Tubes 
    (1D), reduced graphene oxide 
    and MXene 
    (2D). Polydimethylsiloxane 
    , Ecoflex 
    , cotton 
    and polyester 
    are used in resistive tactile sensing applications as matrix components. 
    \item Capacitive Tactile sensors transduce pressure into a change in the capacitance and are typically fabricated by sandwiching a dielectric layer between two parallel electrodes. They typically feature low power consumption, temperature independence, and stability against long-term signal drift at the expense of susceptibility to electromagnetic interference and require a complex measurement circuit. Examples of materials used in electrodes for capacitive tactile sensors are nanometal wires (NWs)
    , indium tin oxide (ITO) 
    , carbon nanotubes (CNTs)
    and graphene 
    . For dielectric layers, typically materials such as polydimethylsiloxane (PDMS) 
    , polyurethane 
    and Ecoflex 
    are used.
    \item Piezoelectric tactile sensors utilize the piezoelectric effect for transduction. The piezoelectric effect causes the deformation of the material to generate voltage due to electric dipole moments. The generated voltage depends on the amount of deformation. These sensors exhibit high sensitivity to dynamic pressure making them great candidates for vibration detection and texture characterization, however, the detection of static pressure is limited because the piezoelectric effect occurs only when the applied stimuli change. Typical examples of piezoelectric materials used of sensor applications are polyvinylidene fluoride (PVDF)
    , zinc oxide(ZnO) 
    and lead zirconate titanate (PZT)
    \item  Triboelectric tactile sensors are a recent addition to this saga which originated from the introduction of triboelectric generator (TENG) in 2012. 
    They feature simple struction and ease of fabrication and scalability. TENGs are fabricated using two materials with different electronegativities covered with electrodes. They generate an electrical potential through contact and separation between them. They are suitable for dynamic pressure sensing, as the output of these sensors depends both on the magnitude and frequency of pressure applied. Various materials such as PDMS
    , polyurethane
    , polytetrafluoroethylene (PTFE)
    , hydrogels
    , ITO 
    and fabrics 
    are used for triboelectric sensors. 
    \item Other niche sensors include: Optical tactile sensors are based on a combination of light emitters and photodetectors and detect the change in intensity or wavelength of light caused by the external pressure on the sensor 
    . They offer high resolution at the expense of integration complexity and high power consumption 
    . Magnetic tactile sensors are based on Hall effect or the principles of giant magnetoresistance 
    to detect the changes (multidirectional) in magnetic flux (magnitude and direction) when external pressure is applied to the sensor. Their performance is prone to environmental noise. Iontronic tactile sensors are a more recent innovation that utilizes the electron double layer or EDL at the interface between the ionic materials and electrode 
    . EDL capacitance due to external deformation is their working principle. Iontronic tactile sensors are an active topic of research for the study of their mechanism and performance improvement potential.  
\end{enumerate}
\red{There are various measures of performance for benchmarking tactile sensors, especially for prosthetic applications. For artificial skin applications, these benchmarking statistics of the sensor composite should resemble the performance of human skin. Some of the affecting parameters include\cite{TouchSensingReview}:}
\begin{itemize}
\color{black}
    \item Spatial resolution is the ability of the sensor to distinguish between two simultaneous touch stimuli. human skin at the fingertips has a spatial resolution of 1\-2 millimeter. \cite{TouchSensingReview20}.
    \item Force sensitivity is defined as the smallest possible detectable contact force variation. Dynamic range is the full range of the sensor's capability. Sensitivity is also affected by the distribution of the applied force over the skin and the range of forces applied. Generally, a sensitivity of 0.01 to 10 Newtons of force and a dynamic range of 1000:1 is enough to resemble human skin capabilities.\cite{TouchSensingReview} 
    \item Linearity of the sensor's output is also desired, as the artificial skin is expected to be able to perform for a range of bandwidth requirements (100Hz for vibrations to 1Hz for discreet touch sensing \cite{TouchSensingReview23}). Typically, these sensors can apply compensation to correct non-linear (e.g. logarithmic) variations.
    \item Frequency response is also a crucial factor, especially for applications like object recognition, grasp stability, pattern recognition, etc. This measurement frequency range varies according to the tactile activity, for example, 10-60 Hz is required for skin stretching compared to 50-100 Hz for vibrations \cite{TouchSensingReview41}. 
    \item Directionality and sensing surface compatibility are essential for tactile sensors in artificial skin applications. Sensors should be compatible with tough and resistant surfaces and be able to sense in multiple directions along normal and tangential planes \cite{TouchSensingReview40}. 
    \item Sensor properties like repeatability, hysteresis, drift, and robustness are crucial. Repeatability is the deviation in sensor output under sinusoidal stimuli. Hysteresis is performance variation during continuous loading and unloading. Drift measures reading variation over time with a constant force. Robustness reflects a sensor's resilience to external factors. Ideally, sensors should exhibit high repeatability, low hysteresis, low drift, and high robustness.
\end{itemize}

Various strategies for enhancing tactile sensor performance, including self-powered sensors, computational techniques, and wireless communication, are discussed in \cite{ReviewFlexTactHIS}. \cite{ProstheticSkin} reviews materials and devices simulating human skin properties. Studies like \cite{CapPiezoTandem}, \cite{PiezoSkin}, and \cite{HandTactileForceFeedback} utilize capacitive and piezoelectric sensors in tactile composites. \cite{PiezoSkin} introduces an SNN Tempotron classifier achieving 99.45\% accuracy in binary textural classification. \cite{PVDFTactile} presents an ARM microcontroller-based interface for eSkin using PVDF-based tactile sensors. Piezoresistive sensors, explored in \cite{DualModePS}, change resistivity upon stimulation. Optimization and new dimensions in tactile sensing methods are actively researched, as seen in \cite{PiezoNanofibre} with a PVDF-TrFE nano-fiber piezoelectric material and \cite{ScreenPrintTactile} employing screen printing on textiles for sensor array implementation. \cite{GaNSelfpowered} covers a gallium nitride-based self-powered tactile sensing array.

\subsection{Limb amputation and Neuroprosthetic approach}

Most of the prosthetic devices that are available commercially do not provide sensory feedback regarding the interaction of the device with its surrounding environment. For a long time now, users have been left to rely on the limited haptic feedback achieved through the interface of the prosthetic socket and residual limb (also known as a stump). Non-invasive solutions for this problem have been demonstrated with limited benefits. These solutions include continuous or time-discrete vibrotactile \cite{ContVibro} \cite{TimeDisVibro}  and electrocutaneous \cite{Electrocutaneous} stimulation. These feedbacks are not homologous, and they only provide an unrefined perception that lacks selectivity of stimulus. \cite{NeuroImpl} Recently, there has been active research in this area to devise a somatosensory feedback system for prosthetic limbs using neural implants. Recently, a kinesthetic prosthetic arm with neural linkages for sensory feedback has proven to provide seamless and lifelike experience for the user \cite{bionicarm2023}.

Traditionally, amputees have required constant visual monitoring of the motor actions of their prosthetic limbs. This routes from the fact that there is a lack of \textit{Proprioception} and \textit{Embodiment} in the user's body regarding the prosthetic limb. Limb amputation also often causes abnormal sensory experiences such as phantom limb syndrome or painful phantom limb pain. Phantom limb syndrome is the perception of a phantom (not real) limb in abnormal anatomical locations. This is also called telescoping \cite{PhantomLimb}. Phantom limb pain refers to the perception of excruciating pain with a knifelike, striking, pricking, or burning sensation at the site of an amputated limb.  \cite{PhantomPain}

Research has shown that intraneural stimulation with visual synchronous feedback has been proven effective in reducing phantom limb sensation while also increasing the embodiment of the prosthetic limb. The technique involved consists of neurotactile stimulation of a nerve location that corresponds to a location on the amputated, limb while synchronously (simultaneously) displaying visualization activity in a virtual simulation of the limb through an HMD. 

Invasive neural implants that are surgically located in specific locations of the concerned nerve can recreate homologous somatotopic sensations. \cite{NeuroImpl} describes a neuroprosthetic leg where neural implants are surgically located onto specific mapped sites of the Tibial nerve. A commercial prosthetic leg with knee and foot was modified with wireless transmission modules in the socket and pressure sensors under the foot. The stimulus collected by the pressure sensors was processed by a portable onboard computer and corresponding signals were wirelessly transmitted to the implanted neuroprosthetic assembly. The demonstrated study had successfully stimulated somatotopic sensation in the phantom limb. The user was able to perceive localized touch under the foot, along with different flexion angles of the prosthetic leg. Experiments conducted with the user demonstrated enhanced ownership of the prosthetic leg by the user during motor activities.\cite{Ch9NPLeg}

\section{Future research}

The development of \textbf{neuroprosthetic hands} will continue to be a significant challenge for improving the lives of individuals with upper-limb amputations. Many amputees will still rely on basic cosmetic or hook-like prosthetic hands due to the enduring complexity of replicating the full range of functions found in natural human hands. This challenge encompasses aspects like appearance, weight, flexibility, dexterity, versatility, and sensory perception. The field of soft robotics is expected to make advancements that will address the existing limitations in grasping force, control speed and precision, and sensory integration. Researchers will explore the development of bioinspired hybrid systems \cite{softrobotics23} that combine soft and rigid materials to improve the functionality of these prosthetic hands. In this regard, 3D printing technology will be a valuable tool for customizing these hybrid systems, but there will be ongoing challenges in terms of material strength and compatibility. 

High-performance soft e-skins will be a focal point, as will the development of portable, compliant feedback stimulation devices that can establish robust connections between neuroprosthetic hands and the user's neural systems. Researchers will strive to create neuroprosthetic hands integrated with multifunctional, high-density e-skins to improve interactions with the external environment, and soft electrodes will be explored to establish myoneural interfaces for natural tactile sensations and better motor control during amputations. Maximisation of the \textbf{temporal resolution} also remains an area of improvement. Researchers will work on improving the real time execution and very compact machine learning models \cite{iqbal2023neural} in special accelerators attached to the sensors. 

\textbf{Symbiotic prostheses} should give personalized help to each user because people with amputations have different physical conditions and movement problems. Right now, in clinics, they personalize robot leg prostheses manually, which takes a lot of time and can be inaccurate. Researchers have made computer programs (like RL algorithms and Bayesian optimization) to make this personalisation process faster and better with the help of humans. These computer programs can adjust the prosthesis in just 5 minutes and keep making it work better over time.

But there are some challenges. The computer systems in these robotic prosthetics need to be more \textbf{reliable and safe} when people use them in their daily lives. And we're not sure how much people with amputations can trust these computer-controlled prostheses. So, in the future, researchers will work on making these computer-powered robotic prostheses more versatile, safe, and acceptable for people with amputations.

\section{Summary and Closing Words}
 This paper is targeted to review some of the recent trends in robotic prosthetics. The paper is primarily focused on understanding the field of soft robotics and the various control strategies that are available for controlling a robotic prosthetic limb. The possibilities that Neuroprosthetics and Brain-Computer Interfaces represent when it comes to providing an alternative to people suffering from various sensory and motor impairments have been explored. While many objects in this discussion of soft robotics and BCI interfaces are currently not ready for commercial application, they are being studied extensively in many research laboratories across the world with the hopes of bringing the next revolution in robotic prosthetics.  This review is ascribed to this field of robotic prostheses that emerges from the intersection of Neuroscience, Mechatronics, Artificial Intelligence, Medicine, and of course surgery.

Soft robotics is an area emerging and evolving at the bleeding edge of chemistry, material science, engineering, and biology. The numerous options in soft actuators, each benefiting from active research and innovation, are representatives of bio-inspired ingenuity. Considering the current readiness of technology, Fluid filled actuators offer huge strains and thereby commendable applicability for the development of robotic prosthetic limbs. Innovations such as Thin McKibben Muscles \cite{ElectricalTMM} in tandem with Electro-hydro-dynamics based flexible fluidic pump \cite{StrPump} and self-healing material \cite{SelfHealReview} are actively contributing to enhancing the desirability and mainstream attention towards FFA actuators.  Electroactive polymers are an interesting variety of soft robotic actuators. The field itself dates back to multiple decades \cite{OrigEAP} but variables of the technology along with potential applications are still subjects of active research. Shape Memory Alloys are already coming into the mainstream of smart material-based mechanical engineering models. Along with their efficient actuator properties, they are also studied for their superelastic properties in applications such as vibration dampening, etc. Electrorheological and Magnetorheological fluids have immense applications in the vibrotactile and force distribution area for prosthetic actuation. Piezoelectric sensors and actuators are being actively used for numerous applications ranging from Brail reading tablets and tactile feedback for robotic surgery, to touch and vibration sensing in various areas. Control strategies are of as much importance for an efficient robotic prosthetic limb as is the actuator itself. TMR, EMG, etc. are established technologies that are already powering many prosthetic limbs.\cite{TMREMG} Advancements in BCI and Integrated AI will surely improve and extend the scope of application for robotic prosthetics to people suffering from worse disabilities. The advancements in sensory rehabilitation through neuroprosthetics with simultaneous development in sensory devices such as artificial skin can potentially integrate the artificial prosthetic with one's subconscious, further empowering the disabled. The future of robotic prosthetics is witnessing blooming potential, as the difference between man and machine becomes fainter with each passing day.

    \bibliographystyle{ACM-Reference-Format}
    \bibliography{ref}

\end{document}